\documentclass[10pt,twocolumn,letterpaper]{article}

\usepackage{cvpr}
\usepackage{times}
\usepackage{epsfig}
\usepackage{graphicx}
\usepackage{amsmath}
\usepackage{amssymb}
\usepackage{soul}
\usepackage{multirow}
\usepackage{csquotes}
\usepackage{rotating}

\cvprfinalcopy % *** Uncomment this line for the final submission

\def\cvprPaperID{2100} % *** Enter the CVPR Paper ID here

\begin{document}

%%%%%%%%% TITLE
\title{Weakly-Supervised Salient Object Detection via Scribble Annotations}

% \and

\author{
Jing Zhang$^{1,3,4}$\quad
Xin Yu$^{1,3,5}$\quad
Aixuan Li$^{^2}$ \quad
Peipei Song$^{1,4}$\quad
Bowen Liu$^{2}$\quad
Yuchao Dai$^2$\thanks{Corresponding author: Yuchao Dai \emph{(daiyuchao@gmail.com)}}\quad\\
$^1$ Australian National University, Australia \quad
$^2$ Northwestern Polytechnical University, China \\
$^3$ ACRV, Australia \quad
$^4$ Data61, Australia \quad 
$^5$ University of Technology Sydney, Australia\\
% {\tt \small \href{http://dpfan.net/UCNet/}{http://dpfan.net/UCNet/}}\\
}

\def\JZ#1{{\color{red}{\bf [Jing:} {\it{#1}}{\bf ]}}}
\def\AX#1{{\color{blue}{\bf [Aixuan:} {\it{#1}}{\bf ]}}}
\def\BW#1{{\color{yellow}{\bf [Bowen:} {\it{#1}}{\bf ]}}}
\def\YD#1{{\color{blue}{\bf [Yuchao:} {\it{#1}}{\bf ]}}}
\def\XY#1{{\color{blue}{\bf [Xin:} {\it{#1}}{\bf ]}}}
\def\PP#1{{\color{red}{\bf [peipei:} {\it{#1}}{\bf ]}}}
\maketitle
%\thispagestyle{empty}

%%%%%%%%% ABSTRACT
\begin{abstract}

Compared with laborious pixel-wise dense labeling, it is much easier to label data by scribbles, which only costs 1$\sim$2 seconds to label one image. However, using scribble labels to learn salient object detection has not been explored. 
In this paper, we propose a weakly-supervised salient object detection model to learn saliency from such annotations. 
In doing so, we first relabel an existing large-scale salient object detection dataset with scribbles, namely S-DUTS dataset.
Since object structure and detail information is not identified by scribbles, directly training with scribble labels will lead to saliency maps of poor boundary localization.
To mitigate this problem, we propose an auxiliary edge detection task to localize object edges explicitly, and a gated structure-aware loss to place constraints on the scope of structure to be recovered.
% to force our predicted saliency maps to align with the image edges. Meanwhile, an auxiliary edge detection task is exploited to localize object edges explicitly, acting as the constraints for our structure loss. 
Moreover, we design a scribble boosting scheme to iteratively consolidate our scribble annotations, which are then employed as
% by applying DenseCRF to our initially estimated saliency map.
% Then, we employ the updated annotations as 
supervision to learn high-quality saliency maps.
As existing saliency evaluation metrics neglect to measure structure alignment of the predictions, the saliency map ranking metric may not comply with human perception.
We present a new metric, termed saliency structure measure, to measure the structure alignment of the predicted saliency maps, which is more consistent with human perception. 
Extensive experiments on six benchmark datasets demonstrate that our method not only outperforms existing weakly-supervised/unsupervised methods, but also is on par with several fully-supervised state-of-the-art models\footnote{Our code and data is publicly available at: \url{https://github.com/JingZhang617/Scribble_Saliency}.}.
\end{abstract}

%%%%%%%%% BODY TEXT
\vspace{-5mm}
\section{Introduction}
\label{sec:intro}

\begin{figure}[!t]
   \begin{center}
   \begin{tabular}{ c@{ } c@{ } c@{ }}
   {\includegraphics[width=0.30\linewidth]{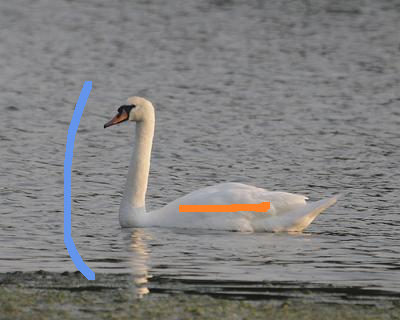}}&
   {\includegraphics[width=0.30\linewidth]{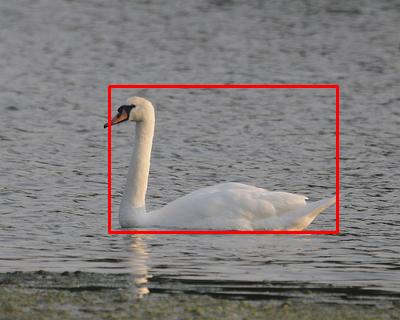}}&
   {\includegraphics[width=0.30\linewidth]{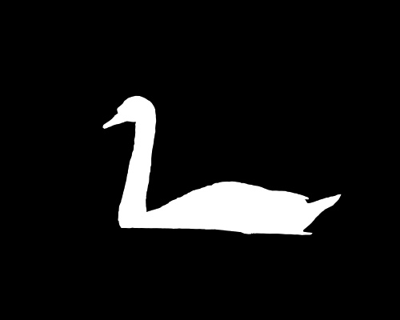}}
   \\
   \footnotesize{(a) GT(scribble)} & \footnotesize{(b) GT(Bbx)} & \footnotesize{(c) GT(per-pixel)}\\
   {\includegraphics[width=0.30\linewidth]{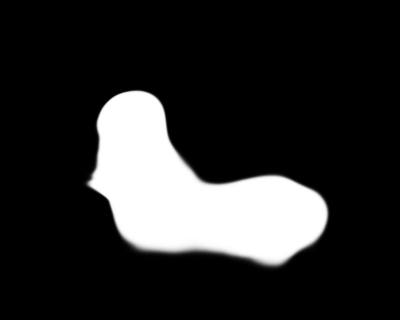}}&
   {\includegraphics[width=0.30\linewidth]{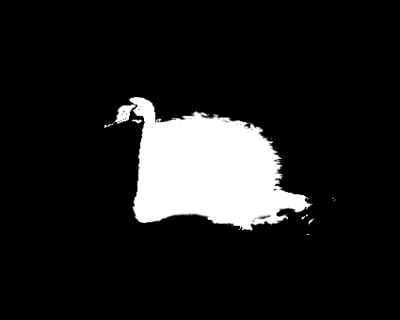}}&
   {\includegraphics[width=0.30\linewidth]{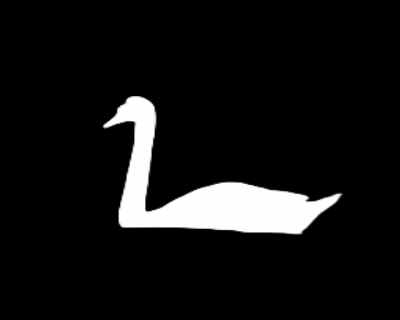}} \\
   \footnotesize{(d) Baseline} & \footnotesize{(e) Bbx-CRF} & \footnotesize{(f) BASNet}\\
   {\includegraphics[width=0.30\linewidth]{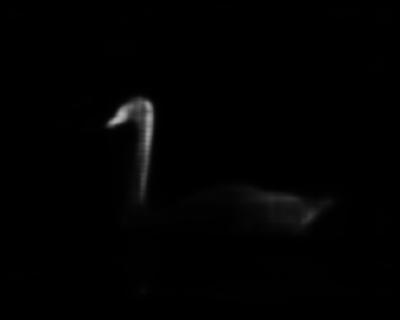}}&
   {\includegraphics[width=0.30\linewidth]{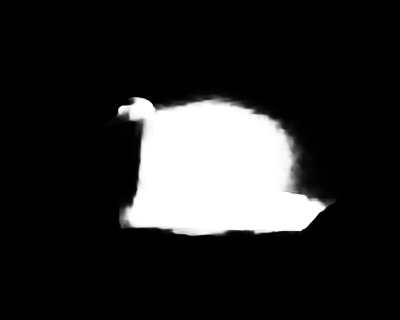}}
   &
   {\includegraphics[width=0.30\linewidth]{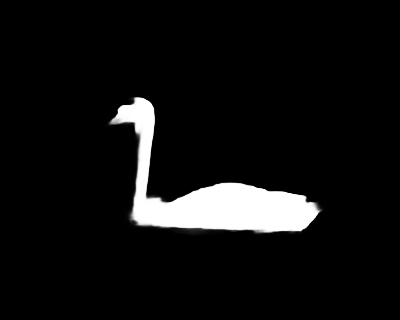}}\\
   \footnotesize{(g) WSS} & \footnotesize{(h) Bbx-Pred} & \footnotesize{(i) Ours}\\
   \end{tabular}
   \end{center}
   \vspace{-2mm}
\caption{\small
% Illustration of our weakly supervised saliency detection from scribble annotations. 
% Scribble annotation (a) is sparse but accurate compared with bounding annotation (b). Directly training with scribble (a) leads to prediction (d) of poor boundary localization. Although bounding box (b) includes more salient region, processing it with DenseCRF \cite{Deeplab} (e) for training will introduce noise to network, and lead to unsatisfied prediction (h). We learn from scribble (a) and obtain (g), which is comparable with
% state-of-the-art fully-supervised method \cite{BASNet_Sal} (f) and much better than competing image-level label based weakly supervised method \cite{imagesaliency} (i).
(a) Our scribble annotations. (b) Ground-truth bounding box. (c) Ground-truth pixel-wise annotations. (d) Baseline model: trained directly on scribbles. (e) Refined bounding box annotation by DenseCRF \cite{Deeplab}. (f) Result of a fully-supervised SOD method \cite{BASNet_Sal}. (g) Result of model trained on image-level annotations \cite{imagesaliency} (h) Model trained on the annotation (e). (i) Our result. 
% \XY{use a snake or sth in this figure}
% (a) the input image; (b) the original per-pixel-wised ground truth; (c) the scribble annotation (orange: salient foreground, blue: background, purple: unknown); (d) saliency map of BASNet \cite{BASNet_Sal}; (e) baseline result of directly training with scribble annotation; (f) result of our proposed method. % saliency prediction (f) 
% comparable with 
}
   \label{fig:figure1}
   \vspace{-4mm}
\end{figure}

% \JZ{Our network should work on both scribble annotation and full annotation. With full annotation provided, we should achieve the best performance.}

Visual salient object detection (SOD) aims at locating interesting regions that attract human attention most in an image. Conventional salient object detection methods \cite{Background-Detection:CVPR-2014,High-Dim-Color-Transform:CVPR-2014} based on hand-crafted features or human experience
% \footnote{We define conventional handcrafted salient object detection models as unsupervised models as no pixel-wise annotations are needed.} 
may fail to obtain high-quality saliency maps in complicated scenarios.
The deep learning based salient object detection models \cite{BASNet_Sal,Zhao_2019_ICCV,SCRN_iccv,jing2020uc} have been widely studied, and significantly boost the saliency detection performance.
% of those conventional saliency models with a large margin,
% Recently, deep learning based salient object detection models \cite{CPD_Sal,BASNet_Sal,Wang_2018_CVPR} have been proposed, significantly boosting the detection performance.\XY{use my version plz we do not need to use which, this makes the sentence more concise.}
However, these methods highly rely on a large amount of labeled data, which require time-consuming and laborious pixel-wise annotations.
To achieve a trade-off between labeling efficiency and model performance, several weakly supervised or unsupervised methods \cite{Guanbin_weaksalAAAI,zeng2019Multi,DeepUSPSDR,Zhang_2018_CVPR} have been proposed to learn saliency from sparse labeled data \cite{Guanbin_weaksalAAAI,zeng2019Multi} or infer the latent saliency from noisy annotations \cite{DeepUSPSDR,Zhang_2018_CVPR}.

% . Those models can be divided into two categories: 1) learn saliency from weak annotations \cite{Guanbin_weaksalAAAI,zeng2019Multi}, including image-level labeling, bounding box, etc and 2) learn saliency from noisy labeling \cite{Zhang_2017_ICCV,Zhang_2018_CVPR}, where they obtain noisy labeling from conventional hand-crafted feature based methods.

In this paper, we propose a new weakly-supervised salient object detection framework by learning from low-cost labeled data, (\ie, scribbles, as seen in Fig.~\ref{fig:figure1}(a)). Here, we opt to scribble annotations because of their flexibility (although bounding box annotation is an option, it's not suitable for labeling winding objects, thus leading to inferior saliency maps, as seen in Fig.~\ref{fig:figure1} (h)). 
%The unlabeled pixels are marked as unknown. 
Since scribble annotations are usually very sparse, object structure and details cannot be easily inferred. Directly training a deep model with sparse scribbles by partial cross-entropy loss \cite{NCut_loss} may lead to saliency maps of poor boundary localization, as illustrated in Fig. \ref{fig:figure1} (d). 
% we need other strategy to preserve the lost boundary information. 

To achieve high-quality saliency maps, we present an auxiliary edge detection network and a gated structure-aware loss to enforce boundaries of our predicted saliency map to align with image edges in the salient region. The edge detection network forces the network to produce feature highlight object structure, and 
% a gated structure-aware loss which enforces the boundaries of our predicted saliency map to align with image edges in the salient region. 
% In particular, we first perform edge detection \cite{ChengEdge} on the training dataset to extract image edges, and employ them as targets for the proposed salient edge-detection branch. 
% Then a gated structure-aware loss is introduced to focus on the salient region, and force the network to produce saliency map with clear structure.
the gated structure-aware loss allows our network to focus on the salient region while ignoring the structure of the background.
% generating saliency maps with clear structure.
We further develop a scribble boosting manner to update our scribble annotations by propagating the labels to larger receptive fields of high confidence. In this way, we can obtain denser annotations as shown in Fig.~\ref{fig:crf_grabcut_ours}~(g).

% the  from a give image as an auxiliary task and then employ our extracted edges as targets in our structure-aware loss function. In this manner, our method predicts saliency maps with clear structure. 
% Two different techniques are proposed to recover the lost detail information: 1) edge detection as auxiliary task to recover edge information; 2) structure-aware loss to penalize prediction not well aligned with edge map of input image.

\begin{figure}[!t]
   \begin{center}
   {\includegraphics[width=0.85\linewidth]{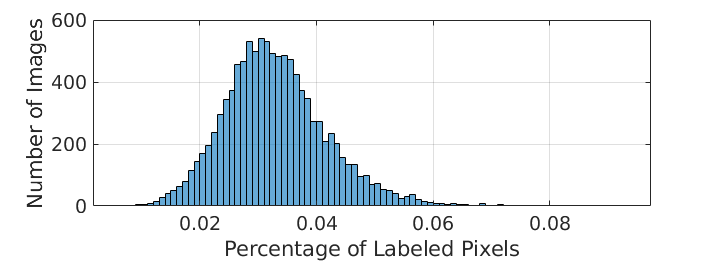}}
   \end{center}
   \vspace{-2mm}
\caption{\small Percentage of labeled pixels in the S-DUTS dataset.}
\vspace{-4mm}
   \label{fig:ratio_scribble}
\end{figure}

Due to the lack of scribble based saliency datasets, we relabel an existing saliency training dataset DUTS \cite{imagesaliency}
% and SALICON \cite{jiang2015salicon} 
with scribbles, namely S-DUTS dataset, to evaluate our method.
% and S-SALICON dataset. 
DUTS is a widely used salient object detection dataset, which contains 10,553 training images.
% and SALICON is the largest eye fixation prediction dataset, which include 10,000 images in the training dataset. 
Annotators are asked to scribble the DUTS dataset according to their first impressions without showing them the ground-truth salient objects.
% (DUTS dataset) or eye fixation ground truth (SALICON dataset). 
% We do not set constraints on size of the scribble annotation.
% and obtain two versions of scribble annotations: 1) scribble based on ground truth and 2) scribble based on RGB image. For the former one, we random select a small part of foreground and background based on the provided ground truth. For the later one, annotators are suggested to label the regions which they define as salient foreground or background. Thus labeling of the first setting is subset of the ground truth, while labeling of the later setting may be different from the ground truth. 
Fig.~\ref{fig:ratio_scribble} indicates the percentage of labeled pixels across the whole S-DUTS dataset. On average, around 3\% of the pixels are labeled (either foreground or background) and the others are left as unknown pixels, demonstrating that the scribble annotations are very sparse.
%which indicates sparsity of the scribble annotations, where around 3\% of the pixels are labeled as foreground or background, with the left 97\% are unknown pixels.

% Furthermore, we extend boundary IOU loss \cite{Luo_2017_CVPR} to boundary IOU measure ($B_\mu$) as a new evaluation metric for saliency models. Specifically, it measures how boundary of the predicted saliency map is aligned with boundary of the ground truth map. We show saliency map of different methods in Fig. \ref{fig:metric_ranking}. The first row represents saliency maps ranking by using mean absolute error (MAE, $\mathcal{M}$), and the second row is based on the boundary iou metric, which shows that $B_\mu$ can reflect sharpness of predictions, while the widely used MAE cannot.

\begin{figure}[!t]
   \begin{center}
   \begin{tabular}{ c@{ } c@{ } c@{ } c@{ } c@{ }}
   {\includegraphics[width=0.165\linewidth]{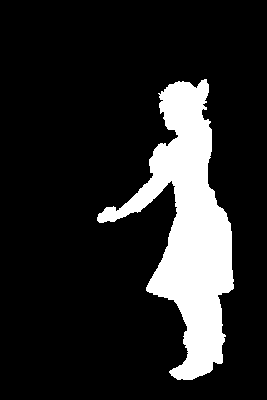}} &
   {\includegraphics[width=0.165\linewidth]{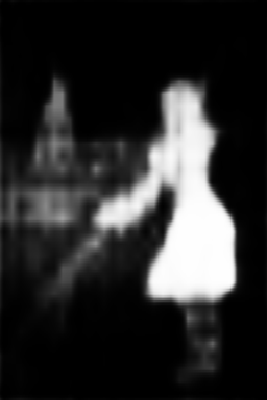}}&
   {\includegraphics[width=0.165\linewidth]{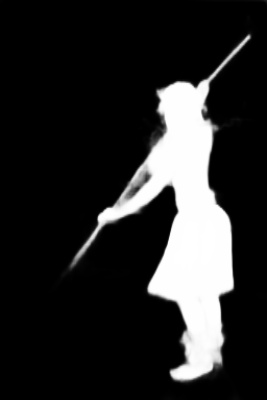}}&
   {\includegraphics[width=0.165\linewidth]{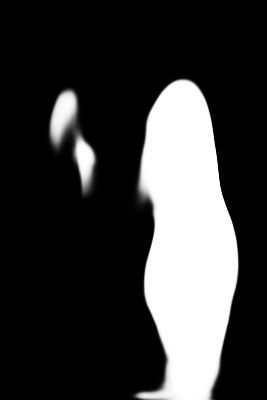}} &
   {\includegraphics[width=0.165\linewidth]{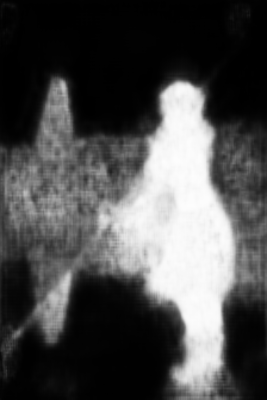}} 
   \\
   \footnotesize{$\mathcal{M}=0$} & \footnotesize{$\mathcal{M}=.054$} & \footnotesize{$\mathcal{M}=.061$}& \footnotesize{$\mathcal{M}=.104$} & \footnotesize{$\mathcal{M}=.144$}\\
   {\includegraphics[width=0.165\linewidth]{metric_illustrate/0001.png}} &
   {\includegraphics[width=0.165\linewidth]{metric_illustrate/0001_MSN.jpg}}&
   {\includegraphics[width=0.165\linewidth]{metric_illustrate/0001_MWS.png}}&
   {\includegraphics[width=0.165\linewidth]{metric_illustrate/0001_SRM.png}} &
   {\includegraphics[width=0.165\linewidth]{metric_illustrate/0001_pce.png}}  \\
   \footnotesize{$\mathcal{B_\mu}=0$} & \footnotesize{$\mathcal{B_\mu}=.356$} & \footnotesize{$\mathcal{B_\mu}=.705$}& \footnotesize{$\mathcal{B_\mu}=.787$} & \footnotesize{$\mathcal{B_\mu}=.890$}\\
   \end{tabular}
   \end{center}
   \vspace{-2mm}
\caption{\small Saliency map ranking based on Mean Absolute Error (1$^{st}$ row) and our proposed Saliency Structure Measure (2$^{nd}$ row).}
   \label{fig:metric_ranking}
   \vspace{-4mm}
\end{figure}

Moreover, the rankings of saliency maps based on traditional mean absolute error (MAE) may not comply with human visual perception. For instance, in the 1$^{st}$ row of Fig.~\ref{fig:metric_ranking}, the last saliency map  is visually better than the fourth one and the third one is better than the second one. We propose saliency structure measure ($B_\mu$) that takes the structure alignment of the saliency map into account. The measurements based on $B_\mu$ are more consistent with human perception, as shown in the 2$^{nd}$ row of Fig. \ref{fig:metric_ranking}.

We summarize our main contributions as: 
(1) we present a new weakly-supervised salient object detection method by learning saliency from scribbles, and introduce a new scribble based saliency dataset S-DUTS;
% and S-SALICON dataset based on SALICON eye fixation prediction dataset \cite{jiang2015salicon} 
(2) we propose a gated structure-aware loss to constrain a predicted saliency map to share similar structure with the input image in the salient region; 
(3) we design a scribble boosting scheme to expand our scribble annotations, thus facilitating high-quality saliency map acquisition;
(4) we present a new evaluation metric to measure the structure alignment of predicted saliency maps, which is more consistent with human visual perception;
(5) experimental results on six salient object detection benchmarks demonstrate that our method outperforms state-of-the-art weakly-supervised algorithms.

% Obtain large amount of pixel-wise labeling is time-consuming and expensive.

% Weakly supervised salient object detection: 1) image-level labeling; 2) noisy labeling; 3) caption labeling

% setting: 1) all scribble dataset
% 2) scribble data with a part of mannual annotations

% ablation study: 
% 1) model naively trained by scribbles.
% 2) sensitivity of different amount of ground truth annotation and pseudo labels

%-------------------------------------------------------------------------
% \vspace{-5mm}

\section{Related Work}
In this section, we briefly discuss related weakly-supervised dense prediction models and approaches to recover detail information from weak annotations.
% (including both salient object detection task and other dense prediction tasks, including semantic segmentation, depth estimation, ...), 

\subsection{Learning Saliency from Weak Annotations}
To avoid requiring accurate pixel-wise labels, salient object detection (SOD) methods attempt to learn saliency from low-cost annotations, such as bounding boxes \cite{6619260}, image-level labels \cite{imagesaliency,Guanbin_weaksalAAAI}, and noisy labels \cite{Zhang_2018_CVPR,Zhang_2017_ICCV, DeepUSPSDR}, \etc. This motivates SOD to be formulated as a weakly-supervised or unsupervised task. 
% \footnote{We define models with low annotation cost are cheap, including image level label, bounding box, caption and etc.}
% In this setting, no pixel-wise annotation is needed, cheaper annotation can lead to cheaper network. 
% Among those models, 
Wang \etal \cite{imagesaliency} introduced a foreground inference network to produce potential saliency maps using image-level labels.
Hsu \etal \cite{HsuBMVC17} presented a category-driven map generator to learn saliency from image-level labels.
Similarly, Li \etal \cite{Guanbin_weaksalAAAI} adopted an iterative learning strategy to update an initial saliency map generated from unsupervised saliency methods by learning saliency from image-level supervision.
% proposed to learn saliency from image-level labels, and 
% started with a set of unsupervised saliency map and define them as initial saliency map. They iteratively train an image classification network with image level labels to obtain class activation map, and updated the above initial saliency map. 
A fully connected CRF \cite{Deeplab} was utilized in \cite{imagesaliency,Guanbin_weaksalAAAI} as a post-processing step to refine the produced saliency map. 
Zeng \etal \cite{zeng2019Multi} 
% claimed that a single weak supervision source usually does not contain enough information to train a well-performing model. Thus,
% they 
proposed to train salient object detection models with diverse weak supervision sources, including category labels, captions, and unlabeled data.
Zhang \etal~\cite{Zhang_2017_ICCV} fused saliency maps from unsupervised methods with heuristics within a deep learning framework. 
% deep model that learns saliency without human annotations, where saliency maps from unsupervised methods are fused with heuristics.
% Very recently, 
In a similar setting, Zhang \etal~\cite{Zhang_2018_CVPR} 
proposed to collaboratively update a saliency prediction module and a noise module to learn a saliency map from multiple noisy labels. 

\subsection{Weakly-Supervised Semantic Segmentation}
% \cite{image_level_pixel_level} proposed to learn object segmentation from image-level label, where the trained model put more weight on important pixels for classification, and they produce segmentation map for testing image by obtaining the smoothed maximum of the multi-channel predictions.
Dai \etal \cite{boxsup} and Khoreva \cite{simple_does_it} proposed to learn semantic segmentation from bounding boxes in a weakly-supervised way. 
% They iteratively update network parameters and pseudo label until reach a pre-defined maximum epoch.
% \cite{simple_does_it} introduced a recursive training strategy to learn semantic segmentation from bounding box, where they iteratively updating the generated training labels from bounding box with two additional object priors.
Hung \etal \cite{Hung_semiseg_2018} randomly interleaved labeled and unlabeled data, and trained a network with an adversarial loss on the unlabeled data for semi-supervised semantic segmentation.
% \cite{multi_evidence_fusion} proposed a curriculum learning pipeline for multi-label object recognition, detection and semantic segmentation, where predictions from multiple tasks are fused by incorporating both metric learning and density-based clustering.
% \cite{pixel_semantic_weak} presented a weak semantic segmentation model by adopting image-level supervision. AffinityNet is proposed to predict semantic affinity between a pair of adjacent image coordinates, thus they propagate discriminative local responses to nearby areas which belong to same semantic entity.
Shi \etal \cite{revisiting_dilated_cnn} tackled the weakly-supervised semantic segmentation problem by using multiple dilated convolutional blocks of different dilation rates to encode dense object localization.
% \cite{weak_semi_sementic} developed Expectation-Maximization (EM) methods for weakly supervised semantic image segmentation under several weakly supervised and semi-supervised settings.
% \cite{8099798} introduced the label-propagation process via random-walk hitting probabilities to learn semantic segmentation with weak annotations, where they propagate the sparse label to unlabeled pixels to generate dense label.
% \cite{weak_instance_class_peak} observed that peaks in a class response map correspond to strong visual cues residing inside each instance, thus they proposed instance segmentation model with image-level labels by exploiting class peak responses.
Li \etal \cite{weak_semantic_iterative_minning} presented an iterative bottom-up and top-down semantic segmentation framework to alternatingly expand object regions and optimize segmentation network with image tag supervision.
Huang \etal \cite{8578831} introduced a seeded region growing technique to learn semantic segmentation with image-level labels. 
Vernaza \etal \cite{8099798} designed a random walk based label propagation method to learn semantic segmentation from sparse annotations.
% They start with discriminative regions and progressively increase the pixel-level supervision by seeded region growing.

\subsection{Recovering Structure from Weak Labels}
% For those weak annotations, as their exists no object structure information, the main solution to achieve dense map with both rich semantic information and accurate detail information is to add extra regularizer to penalize dense map when it's not consistent to image context information. 
As weak annotations do not contain complete semantic region of the specific object, the
% cover the whole object region, 
predicted object structure is often incomplete. To achieve rich and fine-detailed semantic information, additional regularizations are often employed. 
Two main solutions are widely studied, including graph model based methods (\eg CRF \cite{Deeplab}) and boundary based losses \cite{seed_expand_constrain}.
%or boundary based losses \cite{seed_expand_constrain}.
% to encourage prediction with object edge highlighted.
Tang \etal \cite{NCut_loss} introduced a normalized cut loss as a regularizer with partial cross-entropy loss for weakly-supervised image segmentation.
% Wang~\etal~\cite{BPG_scribble} incorporated a boundary prediction branch for weakly supervised semantic segmentation from scribble supervision. Since $\ell_2$ loss was used to regress the boundaries, the final results may suffer from over-smoothness and were not well aligned to image edges.
% Lin~\etal~\cite{scribble_semantic} presented a graphical model based network that propagated information from scribbles to unmarked pixels while learning network parameters.
Tang \etal \cite{Regularized_loss} modeled standard regularizers into a loss function over partial observation for semantic segmentation.
Obukhov~\etal~\cite{gated_crf_loss} proposed a gated CRF loss for weakly-supervised semantic segmentation.
Lampert \etal \cite{seed_expand_constrain} introduced a constrain-to-boundary principle to recover detail information for weakly-supervised image segmentation.

\begin{figure*}[!t]
   \begin{center}
   {\includegraphics[width=0.90\linewidth]{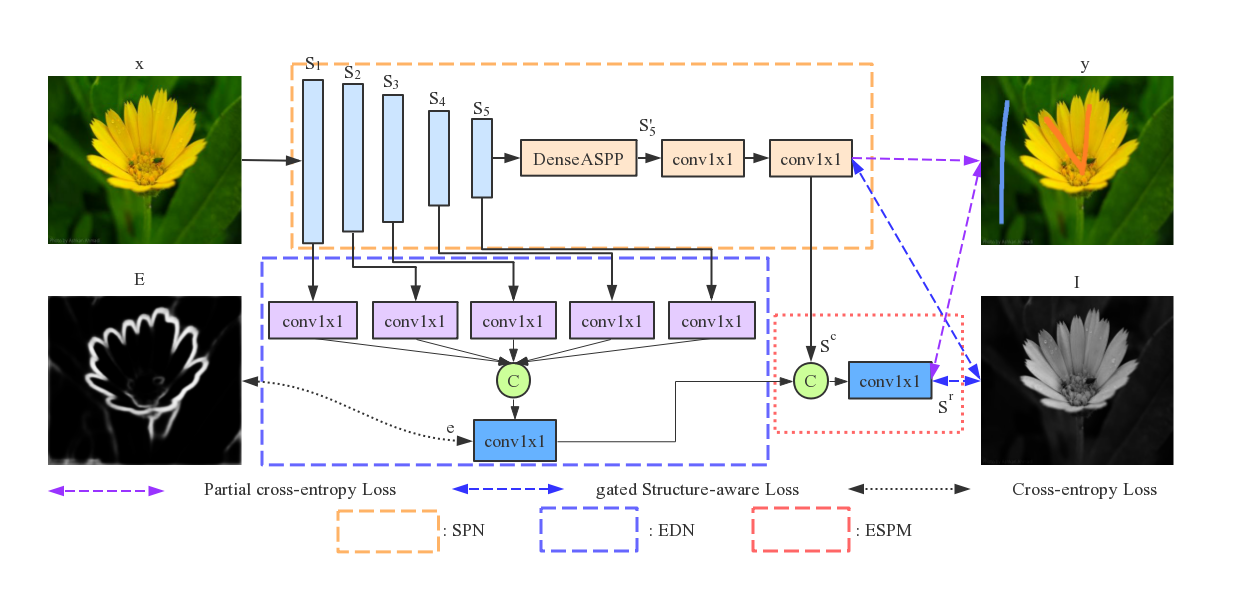}}
   \end{center}
   \vspace{-10mm}
\caption{\small 
% Perceptual illustration of the proposed framework. "CE": cross-entropy loss; "PCE": partial cross-entropy loss \cite{NCut_loss}; "Smoothness loss:" is a boundary-aware loss \cite{occlusion_aware,8100182}; 
% Our network takes RGB image as input, and scribble annotation as supervision signal. Gated Structure-aware loss and the edge detection tasks are introduced to recover object detail information lost in the scribble annotation. \enquote{DenseASPP}: the dense atrous spatial pyramid pooling module from \cite{denseaspp}, shown in Fig. \ref{fig:denseaspp}; 
Illustration of our network. For simplicity, we do not show the scribble boosting mechanism here. \enquote{I} is the intensity image of input \enquote{x}. \enquote{C}: concatenation operation; \enquote{conv1x1}: 1$\times$1 convolutional layer.
}
   \label{fig:overview}
\end{figure*}

% \subsection{Differences from Weakly-Supervised Semantic Segmentation}
\subsection{Comparison with Existing Scribble Models}
Although scribble annotations have been used in weakly-supervised semantic segmentation \cite{scribble_semantic, BPG_scribble}, our proposed scribble based salient object detection method is different from them in the following aspects: 
(1) semantic segmentation methods target at class-specific objects. In this manner, class-specific similarity can be explored.
% for segmentation. 
On the contrary, salient object detection does not focus on class-specific objects, thus object category related
% class-specific similarity 
information is not available.
% in saliency prediction. 
For instance, a leaf can be a salient object while the class category is not available in the widely used image-level label dataset \cite{Imagenet,MSCOCO}. Therefore, we propose edge-guided gated structure-aware loss to obtain structure information from image instead of depending on image category.
% we propose an edge-guided gated structure-aware loss to preserve the structure of our predicted saliency map and a scribble boosting method to refine the prediction; 
(2) although boundary information has been used in \cite{BPG_scribble} to propagate labels, Wang \etal \cite{BPG_scribble} regressed boundaries by an $\ell_2$ loss. Thus, the structure of the segmentation may not be well aligned with the image edges. In contrast, our method minimizes the differences between first order derivatives of saliency maps and images, and leads to saliency map better aligned with image structure.
(3) benefiting from our developed boosting method and the intrinsic property of salient objects, our method requires only scribble on any salient region as shown in Fig.~\ref{fig:diversity_labeling}, while scribbles are required to traverse all those semantic categories for scribble based semantic segmentation \cite{scribble_semantic,BPG_scribble}.

%========================================

%=============================================================
%=============================================================
\section{Learning Saliency from Scribbles}
% \subsection{Problem Formulation}
Let's define our training dataset as: $D=\{x_i,y_i\}_{i=1}^N$, where $x_i$ is an input image, $y_i$ is its corresponding annotation, $N$ is the size of the training dataset. For fully-supervised salient object detection, $y_i$ is a pixel-wise label with 1 representing salient foreground and 0 denoting background.
We define a new weakly-supervised saliency learning problem from scribble annotations, where $y_i$ in our case is scribble annotation used during training, which includes three categories of supervision signal: 1 as foreground, 2 as background and 0 as unknown pixels.
% . Thus, pixels in our scribble label can be divided into three categories: known foreground, known background and unknown pixels.
% , which are highly biased, as indicated 
In Fig. \ref{fig:ratio_scribble}, we show the percentage of annotated pixels of the training dataset, which indicates that around 3\% of pixels are labeled as foreground or background in our scribble annotation.
% In this way, our n: background, foreground and unknown area. We define our setting as weakly supervised saliency detection.

% Given image $x_i$, let $J^s_i$ denotes the labeled pixels, and $J^w_i$ denotes the un-labeled pixels.
%=================
% The conventional way of dealing with weakly-supervised learning problem is adding a regularization term in the loss function:
% \begin{equation}
% \label{semi_super_loss}
%     \mathcal{L} = \mathcal{L}^s + \lambda\mathcal{L}^w,
% \end{equation}
% where $\mathcal{L}^s$ represents the reconstruction error, which usually employs cross-entropy loss for dense prediction tasks, and $\mathcal{L}^w$ is a self-supervised regularizer.
%==================

% In this work, we present a weakly-supervised saliency prediction network with scribble annotations, 
There are three main components in our network,
% and our network mainly consists of three components, 
as illustrated in Fig.~\ref{fig:overview}: 
(1) a saliency prediction network (SPN) to generate a coarse saliency map $s^c$, which is trained on scribble annotations by a partial cross-entropy loss \cite{NCut_loss}; 
(2) an edge detection network (EDN) is proposed to enhance structure of $s^c$, with a gated structure-aware loss employed to force the boundaries of saliency maps to comply with image edges; 
(3) an edge-enhanced saliency prediction module (ESPM) is designed to further refine the saliency maps generated from SPN.% by structure information from the edge detection network.

\subsection{Weakly-Supervised Salient Object Detection}
\textbf{Saliency prediction network (SPN):}
We build our front-end saliency prediction network based on VGG16-Net \cite{VGG} by removing layers after the fifth pooling layer. 
Similar to \cite{HED15}, we group the convolutional layers that generate feature maps of the same resolution as a stage of the network (as shown in Fig. \ref{fig:overview}).
% \XY{what do you mean?}. 
Thus, we denote the front-end model as $f_1(x,\theta)=\{s_1,...,s_5\}$, where $s_m (m=1,...,5)$ represents features from the last convolutional layer in the $m$-th stage (\enquote{relu1\_2, relu2\_2, relu3\_3, relu4\_3, relu5\_3} in this paper), $\theta$ is the front-end network parameters. 

As discussed in \cite{revisiting_dilated_cnn}, enlarging receptive fields by different dilation rates can propagate the discriminative information to non-discriminative object regions. We employ a dense atrous spatial pyramid pooling (DenseASPP) module \cite{denseaspp} on top of the front-end model to generate feature maps $s'_5$ with larger receptive fields from feature $s_5$. In particular, we use varying dilation rates in the convolutional layers of DenseASPP. Then, two extra $1\times1$ convolutional layers are used to map $s'_5$ to a one channel coarse saliency map $s^c$.

As we have unknown category pixels in the scribble annotations, partial cross-entropy loss \cite{NCut_loss} is adopted to train our SPN:
\begin{equation}
    \mathcal{L}_s = \sum_{(u,v)\in J_l}\mathcal{L}_{u,v},
\end{equation}
where $J_l$ represents the labeled pixel set, $(u,v)$ is the pixel coordinates, and $\mathcal{L}_{u,v}$ is the cross-entropy loss at $(u,v)$.

 \begin{figure}[!t]
 \vspace{-10mm}
   \begin{center}
   {\includegraphics[width=0.85\linewidth]{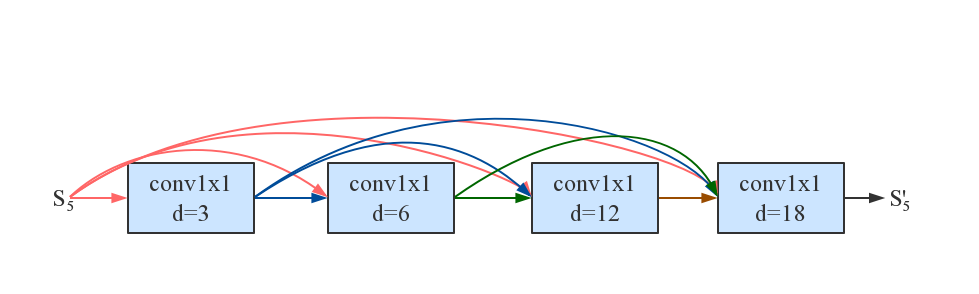}}
   \end{center}
   \vspace{-5mm}
\caption{\small Our \enquote{DenseASPP} module. \enquote{conv1x1 d=3} represents a 1$\times$1 convolutional layer with a dilation rate 3. }
\vspace{-4mm}
   \label{fig:denseaspp}
\end{figure}

% multiple scales of dilated convolution operation can produce network capable of dealing with weak annotations.

\textbf{Edge detection network (EDN):}
Edge detection network encourages SPN to produce saliency features with rich structure information. We use features from the intermediate layers of SPN to produce one channel edge map $e$. Specifically, we map each $s_i (i=1,...,5)$ to a feature map of channel size $M$ with a $1\times1$ convolutional layer. Then we concatenate these five feature maps and feed them to a $1\times1$ convolutional layer to produce an edge map $e$. A cross-entropy loss $\mathcal{L}_e$ is used to train EDN:
\begin{equation}
\label{edge_loss}
    \mathcal{L}_e = \sum_{u,v}(E\log e+(1-E)\log(1-e)),
\end{equation}
where $E$ is pre-computed by an existing edge detector \cite{ChengEdge}.
% an edge map extracted by an existing edge detection network \cite{ChengEdge}.
% We tried three different edge maps as supervision signals, including maps from HED \cite{HED15}, RCF \cite{ChengEdge} and sobel edge from intensity of the image, and achieve the best performance with RCF edge detector.

\textbf{Edge-enhanced saliency prediction module (ESPM):} 
% Inspired by \cite{Ruan2018DevilIT}, 
We introduce an edge-enhanced saliency prediction module to refine the coarse saliency map $s^c$ from SPN and obtain an edge-preserving refined saliency map $s^r$.
% Specifically, we intend to produce rich structure in a saliency map which is also similar to the image.
% with edge map $e$ from the edge detection network. 
Specifically, we concatenate $s^c$ and $e$ and then feed them to a $1\times1$ convolutional layer to produce a saliency map $s^r$. 
Note that, we use the saliency map $s^r$ as the final output of our network.
Similar to training SPN, we employ a partial cross-entropy loss with scribble annotations to supervise $s^r$. 

\textbf{Gated structure-aware loss:}
Although ESPM encourages the network to produce saliency map with rich structure, there exists no constraints on scope of structure to be recovered. Following the \enquote{Constrain-to-boundary} principle \cite{seed_expand_constrain}, we propose a gated structure-aware loss, which encourages the structure of a predicted saliency map to be similar to the salient region of an image.
% edge well-aligned with image edge, no such constraints are applied directly to the produced saliency map $s^c$ or $s^r$. 
% Our basic idea is that the structure of the predict saliency map should be similar to that of the ground truth map. % The basic idea behind smoothness loss is that the predicted saliency map should have consistent value inside an salient objects and sharp distinction happens along object edges. 

We expect the predicted saliency map having consistent intensities inside the salient region and distinct boundaries at the object edges. Inspired by the smoothness loss~\cite{8100182,occlusion_aware}, we also impose such constraint inside the salient regions. Recall that the smoothness loss is developed to enforce smoothness while preserving image structure across the whole image region. However, salient object detection intends to suppress the structure information outside the salient regions. Therefore, enforcing the smoothness loss across the entire image regions will make the saliency prediction ambiguous, as shown in Tabel \ref{tab:ablation_study} \enquote{M3}. 

\begin{figure}[!t]
   \begin{center}
   \begin{tabular}{c@{ } c@{ } c@{ } c@{ } c@{ }}
        {\includegraphics[width=0.184\linewidth]{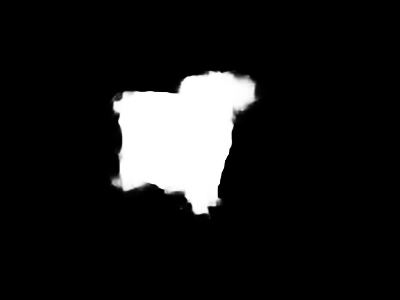}} &
        {\includegraphics[width=0.184\linewidth]{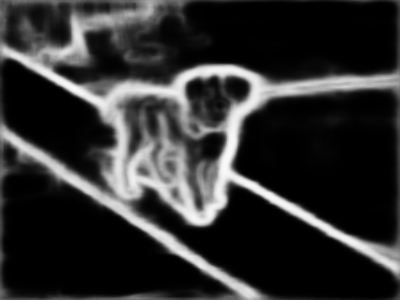}} &
        {\includegraphics[width=0.184\linewidth]{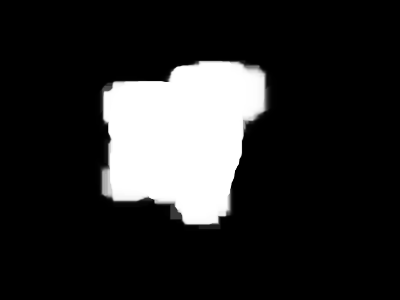}} &
        {\includegraphics[width=0.184\linewidth]{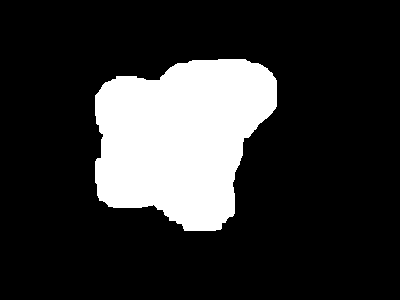}} &
        {\includegraphics[width=0.184\linewidth]{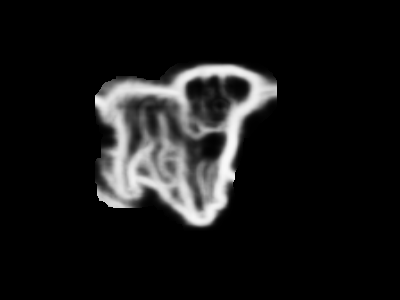}}\\
        \footnotesize{(a)} & \footnotesize{(b)} & \footnotesize{(c)} & \footnotesize{(d)} & \footnotesize{(e)}
   \end{tabular}
   \end{center}
   \vspace{-2mm}
\caption{\small Gated structure-aware constraint: (a) Initial predicted saliency map. (b) Image edge map. (c) Dilated version of (a). (d) Gated mask in Eq. \ref{first_std}. (e) Gated edge map.}
\vspace{-4mm}
   \label{fig:gated_smoothness}
\end{figure}

To mitigate this ambiguity, we employ a gate mechanism to let our network focus on salient regions only to reduce distraction caused by background structure. Specifically, we define the gated structure-aware loss as:
\begin{equation}
\label{first_std}
    \mathcal{L}_b = \sum_{u,v} \sum_{d\in{\overrightarrow{x},\overrightarrow{y}}} \Psi(|\partial_d s_{u,v}|e^{-\alpha |\partial_d (G\cdot I_{u,v})|}),
\end{equation}
% % and the second-order derivative is:
% \begin{equation}
% \label{second_std}
%     dS^2 = \sum_{u,v} \Psi(|\partial^2 s_{u,v}|e^{-\alpha |\partial^2 I_{u,v}|}),
% \end{equation}
where $\Psi$ is defined as $\Psi(s) = \sqrt{s^2+1e^{-6}}$ to avoid calculating the square root of zero, $I_{u,v}$ is the image intensity value at pixel $(u,v)$, $d$ indicates the partial derivatives on the $\overrightarrow{x}$ and $\overrightarrow{y}$ directions, and $G$ is the gate for the structure-aware loss (see Fig .\ref{fig:gated_smoothness} (d)). The gated structure-aware loss applies L1 penalty on gradients of saliency map $s$ to encourages it to be locally smooth, with an edge-aware term $\partial I$ as weight to maintain saliency distinction along image edges.

% We define the gated structure-aware loss in Eq. \ref{first_std} with L1 penalty on gradients of saliency map $s$, which encourages $s$ to be locally smooth. We weight it with an edge-aware term $\partial I$ following \cite{occlusion_aware} to maintain saliency distinction along image edges. Initially, the prediction $s$ is incorrect, and loss $\mathcal{L}_s$ and $\mathcal{L}_b$ in Eq. \ref{final_loss} will be large. As the training progresses, $\mathcal{L}_s$ and $\mathcal{L}_b$ work together to produce better saliency as shown in Fig. (1). Notice that, all the losses in Eq. (4) are used throughout the training. 

Specifically, as shown in Fig. \ref{fig:gated_smoothness}, with predicted saliency map (a)) during training, we dilate it with a square kernel of size $k=11$ to obtain an enlarged foreground region (c)). Then we define gate (d)) as binarized (c)) by adaptive thresholding.
% include the boundaries of salient objects
% In order to include the boundaries of salient objects, we dilate the coarse saliency map with a square kernel of size $k=11$. Fig.~\ref{fig:gated_smoothness} demonstrates the generation procedure of our gate for the structure/smoothness loss. 
As seen in Fig.~\ref{fig:gated_smoothness}(e), our method is able to focus on the saliency region and predict sharp boundaries in a saliency map.

% Then the structure-aware loss is defined as:
% \begin{equation}
% \label{boundary_loss}
%     \mathcal{L}_b = \lambda_1 dS^1+\lambda_2 dS^2.
% \end{equation}

\textbf{Objective Function:} As shown in Fig. \ref{fig:overview}, we employ both partial cross-entropy loss $\mathcal{L}_s$ and gated structure-aware loss $\mathcal{L}_b$ to coarse saliency map $s^c$ and refined map $s^r$, and use cross-entropy loss $\mathcal{L}_e$ for the edge detection network. Our final loss function is then defined as:
\begin{equation}
\label{final_loss}
\begin{aligned}
    \mathcal{L}=&\mathcal{L}_s(s^c,y)+\mathcal{L}_s(s^r,y) \\
    &+\beta_1\cdot\mathcal{L}_b(s^c,x)+\beta_2\cdot\mathcal{L}_b(s^r,x)+\beta_3\cdot\mathcal{L}_e,
\end{aligned}
\end{equation}
where $y$ indicates scribble annotations. The partial cross-entropy loss $\mathcal{L}_s$ takes scribble annotation as supervision, while gated structure-aware loss $\mathcal{L}_b$ leverages image boundary information.
% borrow information from image. 
These two losses do not contradict each other since $\mathcal{L}_s$ focuses on propagating the annotated scribble pixels to the foreground regions (relying on SPN), while $\mathcal{L}_b$ enforces $s^r$ to be well aligned to edges extracted by EDN and prevents the foreground saliency pixels from being propagated to backgrounds.

% With respect to Eq. \eqref{semi_super_loss}, we have:
% % Eq. \ref{semi_super_loss} by defining $\mathcal{L}^s=\mathcal{L}_s(s_c,y)+\mathcal{L}_s(s_r,y)$, and defining $\lambda\mathcal{L}^w=+\beta_1*\mathcal{L}_b(s_c,x)+\beta_2*\mathcal{L}_b(s_r,x)+\beta_3*\mathcal{L}_e$.
% \begin{equation}
% \label{specific_semi_loss}
% \small
%     \left\{
% \begin{array}{lr}
% \mathcal{L}^s=\mathcal{L}_s(s_c,y)+\mathcal{L}_s(s_r,y) & \\
% \lambda\mathcal{L}^w=\beta_1 \mathcal{L}_b(s_c,x)+\beta_2 \mathcal{L}_b(s_r,x)+\beta_3 \mathcal{L}_e.
% \end{array}
% \right.
% \end{equation}

% Then the edge-aware derivative measure is defined as:
% \begin{equation}
% \label{smoothness_loss}
%     eD = \alpha_1 D^1 + \alpha_2 D^2
% \end{equation}

% Then we define the edge-aware saliency measure $E_s$ as:
% \begin{equation}
%     E_s = \frac{2}{1+\exp{(T*eD)}}
% \end{equation}
% where $T$ is a temperature hyperprameter, and we set $T=50$ in this paper.

\subsection{Scribble Boosting}
% With Eq. \ref{final_loss} as our loss function and network structure in Fig. \ref{fig:overview}, we can train a semi-supervised saliency detection model by learning from scribble annotation. As the provided scribble is very sparse, and the ground truth edge map for the edge detection task is obtained from other edge detection model, which is noisy, the resulting saliency map canIn order to produce better saliency mao

% In our weakly supervised saliency model, the $\mathcal{L}^s$ in Eq. \ref{semi_super_loss} is defined as the partial cross-entropy loss, and $\mathcal{L}^w$ represents the prior based regularizer. As supervision signal for loss $\mathcal{L}^s$ is fixed (the scribble annotation), we may failed to discover complex structure quite different with the scribble area. We propose an extension of Eq. \ref{semi_super_loss} as:
% Different from semantic segmentation, salient objects do not exhibit class-specific information. 
While we generate scribbles for a specific image, we simply annotate a very small portion of the foreground and background as shown in Fig. \ref{fig:figure1}. Intra-class discontinuity, such as complex shapes and appearances of objects, may lead our model to be trapped in a local minima, with incomplete salient object segmented.
% when very sparse annotations are used. Thus, methods will fail to uncover entire salient objects.
% Thus, a network may not tell whether a region belonging to foreground or background. For instance, a leaf can be regarded as either saliency object or background. The saliency information only depends on the scribbles.
% Due to the sparsity of our scribble annotations, 
% it difficult to uncover salient objects with complex structure or appearances.
Here, we attempt to propagate the scribble annotations to a denser annotation based on our initial estimation. 

\begin{figure}[!t]
   \begin{center}
   \begin{tabular}{c@{ } c@{ } c@{ } c@{ } }
        {\includegraphics[width=0.22\linewidth]{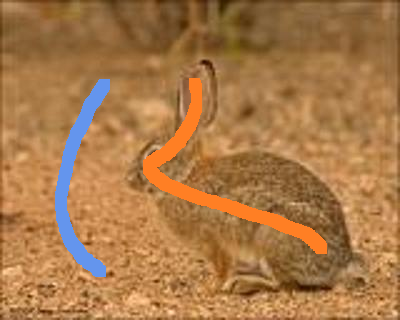}} &
        {\includegraphics[width=0.22\linewidth]{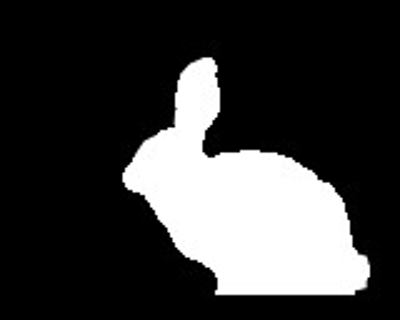}} &
        {\includegraphics[width=0.22\linewidth]{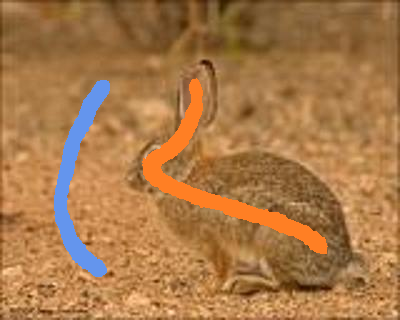}} &
        {\includegraphics[width=0.22\linewidth]{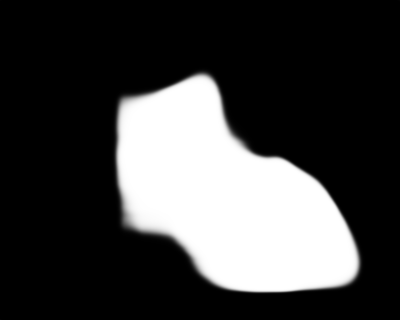}} \\
        % \footnotesize{(a) Scribble} & \footnotesize{(b) $S^c$} & \footnotesize{(c) $CRF(S^c)$} & \footnotesize{(d) New Scribble} & \footnotesize{(e) $S^r$ }\\
        \footnotesize{(a)} & \footnotesize{(b)} & \footnotesize{(c)} & \footnotesize{(d)}\\
        {\includegraphics[width=0.22\linewidth]{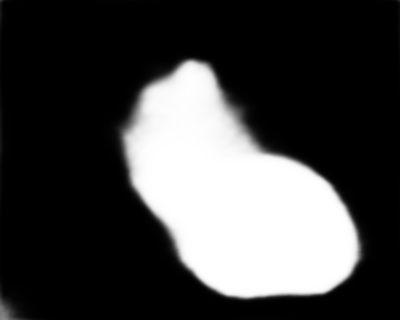}} &
        {\includegraphics[width=0.22\linewidth]{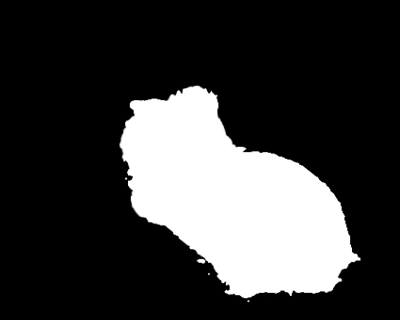}} &
        {\includegraphics[width=0.22\linewidth]{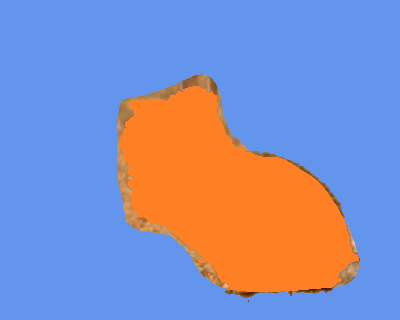}} &
        {\includegraphics[width=0.22\linewidth]{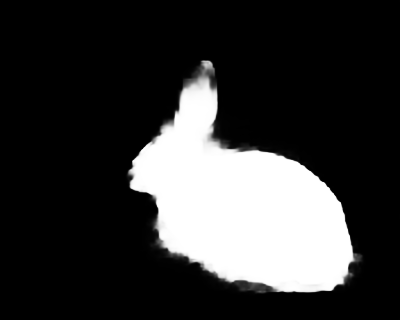}} \\
        \footnotesize{(e)} & \footnotesize{(f)} & \footnotesize{(g)} & \footnotesize{(h)}\\
   \end{tabular}
   \end{center}
   \vspace{-2mm}
\caption{\small 
% Segmentation from CRF, GrabCut, morphological dilation and ours. Trained on scribble (a), we obtain salient foreground segmentation (b). Perform DenseCRF directly on (a) can generate foreground mask (c). By applying GrabCut on raw image with bounding box, we obtain segmentation (d). Expanding scribble annotation in (a) with morphological dilation generates (e).
Illustration of using different strategies to enrich scribble annotations.
(a) Input RGB image and scribble annotations.
(b) Per-pixel wise ground-truth.
(c) Result of applying DenseCRF to scribbles.
(d) Saliency detection, trained on scribbles of (a).
(e) Saliency detection, trained on scribbles of (c).
(f) Applying DenseCRF to the result (d).
(g) The confidence map between (d) and (f) for scribble boosting. Orange indicates consistent foreground, blue represents consistent background, and others are marked as unknown. (h) Our final result trained on new scribble (g).
}
\vspace{-4mm}
   \label{fig:crf_grabcut_ours}
\end{figure}

% \begin{figure}[!t]
%   \begin{center}
%   \begin{tabular}{c@{ } c@{ } c@{ } c@{ }  c@{ }}
%         {\includegraphics[width=0.184\linewidth]{crf_grabcut_show/ILSVRC2012_test_00000019.png}} &
%         {\includegraphics[width=0.184\linewidth]{crf_grabcut_show/ILSVRC2012_test_00000019_our_seg.png}} &
%         {\includegraphics[width=0.184\linewidth]{crf_grabcut_show/ILSVRC2012_test_00000019_crf_seg.png}} &
%         {\includegraphics[width=0.184\linewidth]{crf_grabcut_show/ILSVRC2012_test_00000019_grabcut.png}} &
%         {\includegraphics[width=0.184\linewidth]{crf_grabcut_show/ILSVRC2012_test_00000019_scribble_dilate.png}}\\
%         % \footnotesize{(a) Scribble} & \footnotesize{(b) $S^c$} & \footnotesize{(c) $CRF(S^c)$} & \footnotesize{(d) New Scribble} & \footnotesize{(e) $S^r$ }\\
%         \footnotesize{(a)} & \footnotesize{(b)} & \footnotesize{(c)} & \footnotesize{(d)} & \footnotesize{(e)}\\
%   \end{tabular}
%   \end{center}
%   \vspace{-2mm}
% \caption{\small Segmentation from CRF, GrabCut, morphological dilation and ours. Trained on scribble (a), we obtain salient foreground segmentation (b). Perform DenseCRF directly on (a) can generate foreground mask (c). By applying GrabCut on raw image with bounding box, we obtain segmentation (d). Expanding scribble annotation in (a) with morphological dilation generates (e).}
%   \label{fig:crf_grabcut_ours}
% \end{figure}

% Towards this goal, w
% We aim at
% % designing a scribble annotation boosting mechanism to 
% producing denser and expanded annotations for salient foreground regions. 
A straightforward solution to obtain denser annotations is to expand scribble labels by using DenseCRF \cite{Deeplab}, as shown in Fig. \ref{fig:crf_grabcut_ours}(c). However, as our scribble annotations are very sparse, DenseCRF fails to generate denser annotation from our scribbles
% directly applying DenseCRF on the sparse annotations does not yield much denser annotations 
(see Fig. \ref{fig:crf_grabcut_ours}(c)). As seen in Fig. \ref{fig:crf_grabcut_ours}(e), the predicted saliency map trained on (c) is still very similar to the one supervised by original scribbles (see Fig. \ref{fig:crf_grabcut_ours}(d)).

% , but we did not see significant improvement using this solution due to the sparsity of scribbles. In 
% \XY{give an example in this case plz.}. 
Instead of expanding the scribble annotation directly, we apply DenseCRF to our initial saliency prediction $s^{\text{init}}$, and update $s^{\text{init}}$ to $s^{\text{crf}}$.
% Then, we re-train our network with new scribble based on the updated annotations $s^{\text{crf}}$ and initial map $s^{\text{init}}$. 
Directly training a network with $s^{\text{crf}}$ will introduce noise to the network as $s^{\text{crf}}$ is not the exact ground-truth. We compute difference of $s^{\text{init}}$ and $s^{\text{crf}}$, and define pixels with $s^{\text{init}}=s^{\text{crf}}=1$ as foreground pixels in the new scribble annotation, $s^{\text{init}}=s^{\text{crf}}=0$ as background pixels, and others as unknown pixels.
% Since DenseCRF takes image structure information into account to update $s^{\text{init}}$, $s^{\text{crf}}$ also aligns to input images.
In Fig. \ref{fig:crf_grabcut_ours} (g) and Fig. \ref{fig:crf_grabcut_ours} (h), we illustrate the intermediate results of scribble boosting.
Note that, our method achieves better saliency prediction results than the case of applying DenseCRF to the initial prediction (see Fig. \ref{fig:crf_grabcut_ours} (f)). This demonstrates the effectiveness of our scribble boosting scheme. 
In our experiments, after conducting one iteration of our scribble boosting step, our performance is almost on par with fully-supervised methods.

% With the trained model $f(\omega)$ ($\omega$ is the network parameter set) and given training image set $\{x_i\}_{i=1}^N$, we compute saliency prediction $s=f(x;\omega)$ of the training image set. Then DenseCRF \cite{Deeplab} is performed on $f(x;\omega)$ to produce sharp saliency map $s^{crf}$. Directly training a network with $s^{crf}$ may introduce noise to the network as $s^{crf}$ may not be the exact ground truth. We then compute difference of $s$ and $s^{crf}$, and define pixels with $s=s^{crf}=1$ as foreground pixels in the new scribble annotation, $s=s^{crf}=0$ as background pixels, and $s!=s^{crf}$ as unknown pixels. 

\subsection{Saliency Structure Measure}

% \textbf{Gated CRF Loss}

% \subsection{Edge-aware Mean Absolute Error}

Existing saliency evaluation metrics (Mean Abosolute Error, Precision-recall curves, F-measure, E-measure \cite{Fan2018Enhanced} and S-measure \cite{fan2017structure}) focus on measuring accuracy of the prediction,
% saliency map, 
while neglect whether a predicted saliency map complies with human perception or not. In other words, the estimated saliency map should be aligned with object structure of the input image. In \cite{Luo_2017_CVPR}, bIOU loss was proposed to penalize on saliency boundary length. We adapt the bIOU loss as an error metric $B_\mu$ to evaluate the structure alignment between saliency maps and their ground-truth.

% sharpness of the prediction and present a saliency structure measure $B_\mu$, which
% % the effectiveness of how those loss works 
% is complementary to existing metrics. As shown in Fig.3, our new metric is more consistent with human visual perception.
% We thus present a saliency structure measure $B_\mu$, motivated by the boundary IOU loss \cite{Luo_2017_CVPR}, to evaluate the structure alignment between saliency maps and their ground-truth. 

% Since edges represent image structure, we compare the binarized edge maps between saliency maps and images as our our saliency structure measure. To achieve an edge map, we binarize the gradient magnitude image. Specifically, when the gradient magnitude is larger than 0\XY{a threshold?}, we set it to 1.

\begin{figure}[!t]
   \begin{center}
   {\includegraphics[width=0.24\linewidth]{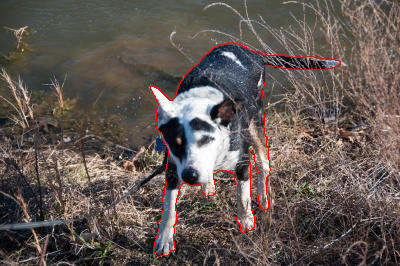}}
   {\includegraphics[width=0.24\linewidth]{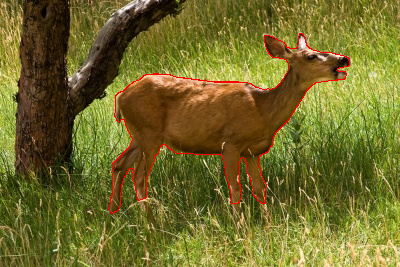}}
   {\includegraphics[width=0.24\linewidth]{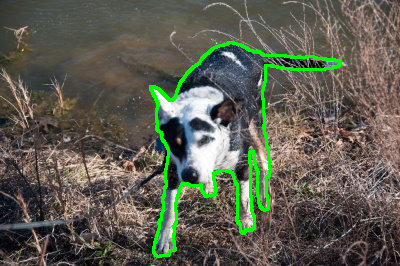}}
   {\includegraphics[width=0.24\linewidth]{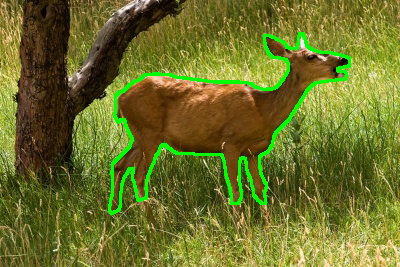}}
   \end{center}
   \vspace{-2mm}
\caption{\small
% Edge map dilation for $B_\mu$ measure. 
The first two images show the original image edges. We dilate the original edges (last two images) to avoid misalignments due to the small scales of original edges.
% , they are sensitive to misalignments. The last two images show the dilated edges.
}
   \label{fig:re-align_edge_map}
   \vspace{-4mm}
\end{figure}

Given a predicted saliency map $s$, and its pixel-wise  ground truth $y$,
% \footnote{In the evaluation stage, $y$ represents per-pixel wise labeling.}', 
their binarized edge maps are defined as $g_s$ and $g_y$ respectively.
Then $B_\mu$ is expressed as: $B_\mu = 1-\frac{2\cdot\sum(g_s\cdot g_y)}{\sum(g_s^2+g_y^2)}$,
% \begin{equation}
%     B_\mu = 1-\frac{2\cdot\sum(g_s\cdot g_y)}{\sum(g_s^2+g_y^2)},
% \end{equation}
where $B_\mu \in [0,1]$. $B_\mu=0$ represents perfect prediction. 
As edges of prediction and ground-truth saliency maps may not be aligned well due to the small scales of edges, they will lead to unstable measurements (see Fig. \ref{fig:re-align_edge_map}). We dilate both edge maps with square kernel of size 3 before we compute the $B_\mu$ measure.
As shown in Fig. \ref{fig:metric_ranking}, $B_\mu$ reflects the sharpness of predictions which is consistent with human perception.
% and produces the highest value (worst) for the smoothest prediction (last image in the second row).

\subsection{Network Details}
We use VGG16-Net \cite{VGG} as our backbone network. In the edge detection network, we encode $s_m$ to feature maps of channel size 32 through $1\times1$ convolutional layers. In the \enquote{DenseASPP} module (Fig. \ref{fig:denseaspp}), the first three convolutional layers produce saliency features of channel size 32, and the last convolutional layer map the feature maps to $s'_5$ of same size as $s_5$. Then we use two sequential convolutional layers to map $s'_5$ to
% A $1\times 1$ convolutional layer is used to map $s'_5$ to feature maps of $32$ channels, and then we obtain 
one channel coarse saliency map $s^c$.
% by another $1\times1$ convolutional layer. 
The hyper-parameters in Eq. \ref{first_std} and Eq. \eqref{final_loss} are set as: $\alpha=10$, $\beta_1=\beta_2=0.3$, $\beta_3=1$.

We train our model for 50 epochs. The saliency prediction network is initialized with parameters from VGG16-Net \cite{VGG} pretrained on ImageNet \cite{Imagenet}. The other newly added convolutional layers are randomly initialized with $\mathcal{N}(0,0.01)$. The base learning rate is initialized as 1e-4. The whole training takes 6 hours with a training batch size 15 on a PC with a NVIDIA GeForce RTX 2080 GPU.

% ==============================
\section{Experimental Results}
\label{sec:experimental_results}

\subsection{Scribble Dataset}
\begin{figure}[!t]
  \begin{center}
  \begin{tabular}{ c@{ } c@{ } c@{ } c@{ } c@{ }}
  {\includegraphics[width=0.180\linewidth]{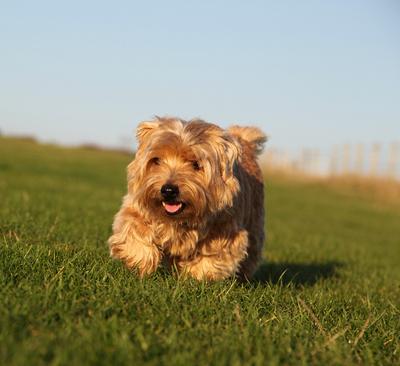}} &
  {\includegraphics[width=0.180\linewidth]{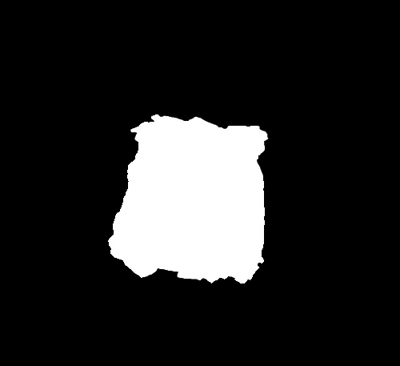}}&
  {\includegraphics[width=0.180\linewidth]{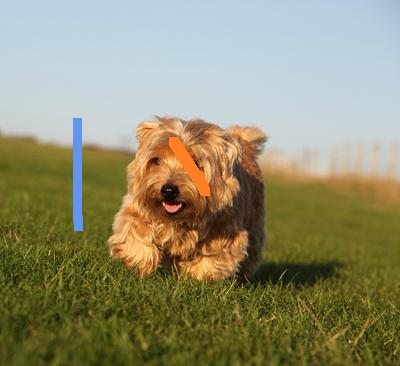}}&
  {\includegraphics[width=0.180\linewidth]{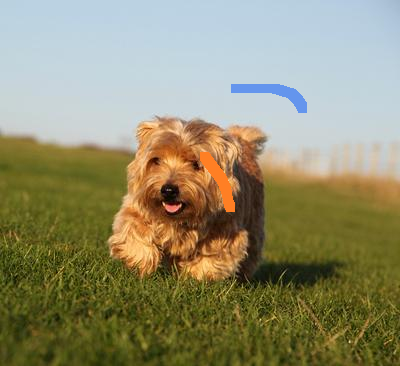}} &
  {\includegraphics[width=0.180\linewidth]{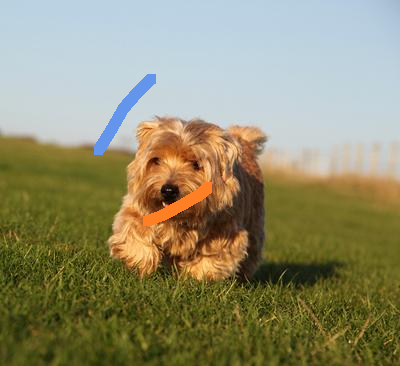}} 
  \\
  {\includegraphics[width=0.180\linewidth]{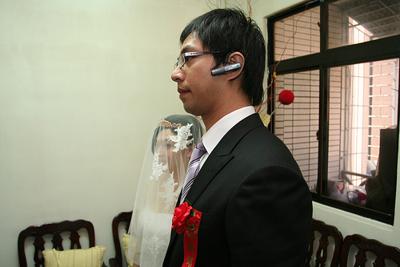}} &
  {\includegraphics[width=0.180\linewidth]{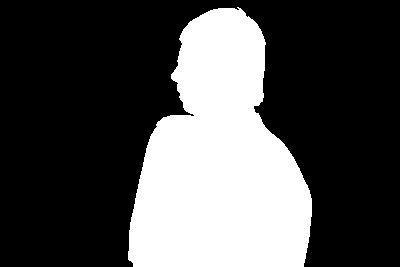}}&
  {\includegraphics[width=0.180\linewidth]{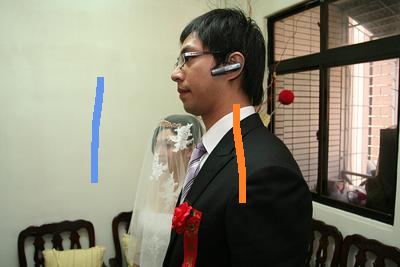}}&
  {\includegraphics[width=0.180\linewidth]{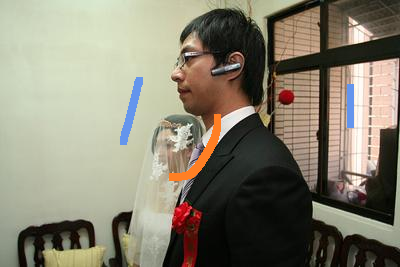}} &
  {\includegraphics[width=0.180\linewidth]{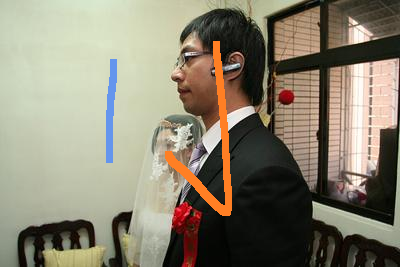}}
%   {\includegraphics[width=0.185\linewidth]{different_annotations/COCO_train2014_000000000307.jpg}} &
%   {\includegraphics[width=0.185\linewidth]{different_annotations/COCO_train2014_000000000307.png}}&
%   {\includegraphics[width=0.185\linewidth]{different_annotations/COCO_train2014_000000000307_s1.png}}&
%   {\includegraphics[width=0.185\linewidth]{different_annotations/COCO_train2014_000000000307_s2.png}} &
%   {\includegraphics[width=0.185\linewidth]{different_annotations/COCO_train2014_000000000307_s3.png}} \\
%   {\includegraphics[width=0.185\linewidth]{different_annotations/COCO_train2014_000000000450.jpg}} &
%   {\includegraphics[width=0.185\linewidth]{different_annotations/COCO_train2014_000000000450.png}}&
%   {\includegraphics[width=0.185\linewidth]{different_annotations/COCO_train2014_000000000450_s1.png}}&
%   {\includegraphics[width=0.185\linewidth]{different_annotations/COCO_train2014_000000000450_s2.png}} &
%   {\includegraphics[width=0.185\linewidth]{different_annotations/COCO_train2014_000000000450_s3.png}}
  \end{tabular}
  \end{center}
  \vspace{-2mm}
\caption{\small Illustration of scribble annotations by different labelers.
% Images of the first two rows come from the S-DUTS dataset, and the left two images are from S-SALICON dataset. 
From left to right: input RGB images, pixel-wise ground-truth labels, scribble annotations by three different labelers.}
  \label{fig:diversity_labeling}
  \vspace{-4mm}
\end{figure}

In order to evaluate our weakly-supervised salient object detection method, we train our network on our S-DUTS dataset labeled by three annotators.
% \footnote{We will release our S-DUTS dataset as well as our training codes for research reproducibility.}.
% We will release three versions of S-DUTS dataset labeled by three different annotators.
In Fig. \ref{fig:diversity_labeling}, we show two examples of scribble annotations by different labelers.
Due to the sparsity of scribbles, the annotated scribbles do not have large overlaps. Thus, majority voting is not conducted.
% As we directly label on raw RGB image, the produced scribble may not be consistent with the provided per-pixel wise GT as shown in second row of Fig. \ref{fig:diversity_labeling}.
As aforementioned, labeling one image with scribbles is very fast, which only takes 1$\sim$2 seconds on average.
%with scribble, and 3$\sim$4 days to label the whole S-DUTS dataset.
% \footnote{We have three annotators to label both of the two scribble dataset.}.
% and do not perform majority voting as conventional per-pixel wise ground truth saliency map generation mechanism, 
% SALICON dataset capture human eye fixation with mouse clicking on the fixation points. The dense eye fixation map is generated by performing Gaussian blur on the fixation points map. The relabeled S-SALICON dataset can reflect both intensity of fixation as well as salient object information, which can be used in the joint saliency and fixation learning framework.
% , which can be used in both eye fixation prediction task and salient object detection task.

\begin{table*}[t!]
  \centering
  \scriptsize
  \renewcommand{\arraystretch}{1.0}
  \renewcommand{\tabcolsep}{1.0mm}
  \caption{Evaluation results
%   of sixteen leading deep RGB salient object detection models 
  on six benchmark datasets.
  $\uparrow \& \downarrow$ denote larger and smaller is better, respectively.
  }\label{tab:BenchmarkResults}
  \begin{tabular}{lr|ccccccccccc|cccccc}
  \hline
%   \toprule
  &  &\multicolumn{11}{c|}{Fully Sup. Models}&\multicolumn{6}{c}{Weakly Sup./Unsup. Models} \\
    & Metric &
   DGRL  & UCF  & PiCANet & R3Net    & NLDF &
   MSNet & CPD & AFNet & PFAN  & PAGRN& BASNet & SBF & WSI & WSS & MNL & MSW & Ours\\
  &  & \cite{Wang_2018_CVPR}        & \cite{UCF_ICCV}       & \cite{picanet}          & \cite{R3Net}              & \cite{Luo_2017_CVPR} &
        \cite{MSNet_Sal}   & \cite{CPD_Sal}                 & \cite{AFNet_Sal} 
        & \cite{pyramid_attention}   & \cite{prpgressive_attention} &\cite{BASNet_Sal}       & \cite{Zhang_2017_ICCV} &   \cite{Guanbin_weaksalAAAI}   & \cite{imagesaliency} & \cite{Zhang_2018_CVPR}     & \cite{zeng2019Multi} & \\
  \hline
  \multirow{4}{*}{\textit{ECSSD}}
    & $B_{\mu}\downarrow$    & .4997 & .6990 & .5917 & .4718 & .5942 & .5421 & .4338 & .5100 & .6601 & .5742 & \textbf{.3642} & .7587 & .8007 & .8079& .6806 & .8510 & \textbf{.5500} \\
    & $F_{\beta}\uparrow$     & .9027 & .8446 & .8715 & \textbf{.9144} & .8709 & .8856 & .9076 & .9008 & .8592 & .8718 & .9128 & .7823 & .7621 & .7672& .8098 & .7606& \textbf{.8650}   \\
    & $E_{\xi}\uparrow$       & .9371 & .8870 & .9085 & \textbf{.9396} & .8952 & .9218 & .9321 & .9294 & .8636 & .8869 & .9378 & .8354 & .7921 & .7963& .8357 & .7876& \textbf{.9077}  \\
    & $\mathcal{M}\downarrow$ & .0430 & .0705 & .0543 & .0421 & .0656 & .0479 & .0434 & .0450 & .0467 & .0644 & \textbf{.0399} & .0955 & .0681 & .1081 & .0902 & .0980& \textbf{.0610}  \\ 
    \hline
     \multirow{4}{*}{\textit{DUT}}
     & $B_{\mu}\downarrow$    & .6188 & .8115 & .6846 & .6061 & .7148 & .6415 & .5491 & .6027 & .6443 & .6447 & \textbf{.4803} & .8119 & .8392 & .8298& .7759 & .8903 & \textbf{.6551} \\
    & $F_{\beta}\uparrow$     & .7264 & .6318 & .7105 & .7471 & .6825 & .7095 & .7385 & .7425 & .7009 & .6754 & \textbf{.7668} & .6120 & .6408 & .5895 & .5966 & .5970& \textbf{.7015}  \\
    & $E_{\xi}\uparrow$       & .8446 & .7597 & .8231 & .8527 & .7983 & .8306 & .8450 & .8456 & .7990 & .7717 & \textbf{.8649} & .7633 & .7605 & .7292 & .7124 & .7283& \textbf{.8345}  \\
    & $\mathcal{M}\downarrow$ & .0632 & .1204 & .0722 & .0625 & .0796 & .0636 & .0567 & .0574 & .0615 & .0709 & \textbf{.0565} & .1076 & .0999 & .1102 & .1028 & .1087& \textbf{.0684} \\ 
   \hline
    \multirow{4}{*}{\textit{PASCAL-S}}
    & $B_{\mu}\downarrow$    & .6479 & .7832 & .7037 & .6623 & .7313 & .6708 & .6162 & .6586 & .7097 & .6915 & \textbf{.5819} & .8146 & .8550 & .8309& .7762 & .8703 & \textbf{.6648} \\
    & $F_{\beta}\uparrow$     & \textbf{.8289} & .7873 & .7985 & .7974 & .7933 & .8129 & .8220 & .8241 & .7544 & .7656 & .8212 & .7351 & .6532 & .6975 & .7476 & .6850& \textbf{.7884}  \\
    & $E_{\xi}\uparrow$       & \textbf{.8353}& .7953 & .8045 & .7806 & .7828 & .8219 & .8197 & .8269 & .7464 & .7545 & .8214 & .7459 & .6474 & .6904 & .7408 & .6932& \textbf{.7975}  \\
    & $\mathcal{M}\downarrow$ & \textbf{.1150} & .1402 & .1284 & .1452 & .1454 & .1193 & .1215 & .1155 & .1372 & .1516 & .1217 & .1669 & .2055 & .1843 & .1576 & .1780&\textbf{.1399}  \\ 
    \hline
     \multirow{4}{*}{\textit{HKU-IS}}
     & $B_{\mu}\downarrow$    & .4962 & .6788 & .5608 & .4765 & .5525 & .4979 & .4211 & .4828 & .5302 & .5329 & \textbf{.3593} & .7336 & .7824 & .7517& .6265 & .8295 & \textbf{.5369} \\ 
     & $F_{\beta}\uparrow$ & .8844 & .8189 & .8543 & .8923 & .8711 & .8780 & .8948 & .8877 & .8717 & .8638 & \textbf{.9025} & .7825 & .7625 & .7734 & .8196 & .7337 &\textbf{.8576}  \\
    & $E_{\xi}\uparrow$       & .9388 & .8860 & .9097 & .9393 & .9139 & .9304 & .9402 & .9344 & .8982 & .8979 & \textbf{.9432} & .8549 & .7995 & .8185 & .8579 & .7862 & \textbf{.9232} \\
    & $\mathcal{M}\downarrow$ & .0374 & .0620 & .0464 & .0357 & .0477 & .0387 & .0333 & .0358 & .0424 & .0475 & \textbf{.0322} & .0753 & .0885 & .0787 & .0650 & .0843 &\textbf{.0470} \\ 
    \hline
     \multirow{4}{*}{\textit{THUR}}
     & $B_{\mu}\downarrow$    & .5781 & - & .6589 & - & .6517 & .6196 & .5244 & .5740 & .7426 & .6312 & \textbf{.4891} & .7852 & - & .7880& .7173 & - & \textbf{.5964} \\
    & $F_{\beta}\uparrow$     & .7271 & . & .7098 & - & .7111 & .7177 & \textbf{.7498} & .7327 & .6833 & .7395 & .7366 & .6269 & - & .6526 & .6911 & -& \textbf{.7181} \\
    & $E_{\xi}\uparrow$       & .8378 & . & .8211 & - & .8266 & .8288 & \textbf{.8514} & .8398 & .8038 & .8417& .8408 & .7699 & - & .7747 & .8073 & - & \textbf{.8367}\\
    & $\mathcal{M}\downarrow$ & .0774 & . & .0836 & - & .0805 & .0794 & .0935 & .0724 & .0939 & \textbf{.0704}& .0734  & .1071 & - & .0966 & .0860 & - &\textbf{.0772} \\
    \hline
  \multirow{4}{*}{\textit{DUTS}}
     & $B_{\mu}\downarrow$    & .5644 & .7956 & .6348 & - & .6494 & .5823 & .4618 & .5395 & .6173 & .5870 & \textbf{.4000} & .8082 & .8785 & .7802& .7117 & .8293 & \textbf{.6026} \\
    & $F_{\beta}\uparrow$     & .7898 & .6631 & .7565 & - & .7567 & .7917 & \textbf{.8246} & .8123 &  .7648 & .7781 & .8226 & .6223 & .5687 & .6330& .7249 & .6479& \textbf{.7467}   \\
    & $E_{\xi}\uparrow$       & .8873 & .7750 & .8529 & - & .8511 & .8829 & \textbf{.9021} & .8928 & .8301 & .8422 & .8955 & .7629 & .6900 & .8061 & .8525 & .7419& \textbf{.8649}  \\
    & $\mathcal{M}\downarrow$ & .0512 & .1122 & .0621 & - & .0652 & .0490 & \textbf{.0428} & .0457 & .0609& .0555 & .0476 & .1069 & .1156 & .1000 & .0749 & .0912 & \textbf{.0622} \\
    % \midrule
    % \bottomrule
  \hline
  \end{tabular}
  \label{tab:deep_unsuper_Performance_Comparison}
\end{table*}

\begin{figure*}[!t]
   \begin{center}
   \begin{tabular}{ c@{ } c@{ } c@{ } c@{ }  c@{ }  c@{ } c@{ } c@{ } c@{ } c@{ }}
   {\includegraphics[width=0.090\linewidth]{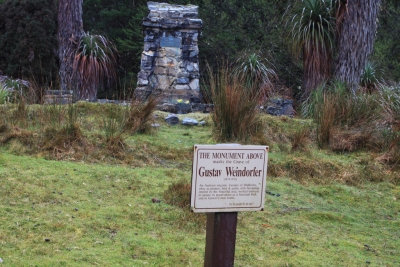}}&
   {\includegraphics[width=0.090\linewidth]{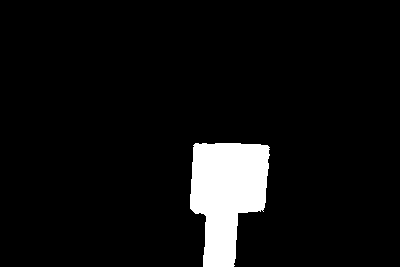}}&
   {\includegraphics[width=0.090\linewidth]{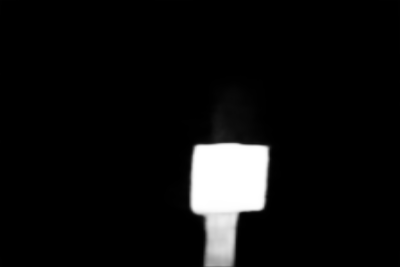}}&
   {\includegraphics[width=0.090\linewidth]{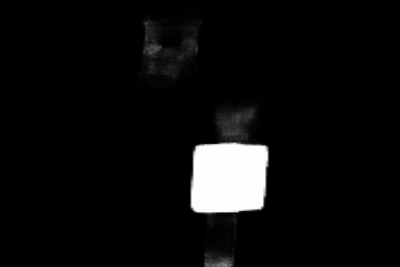}}&
   {\includegraphics[width=0.090\linewidth]{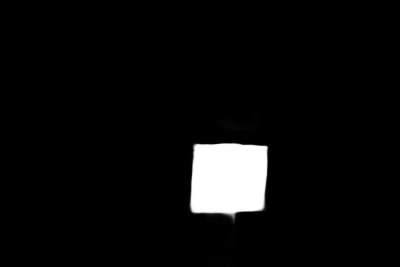}}&
   {\includegraphics[width=0.090\linewidth]{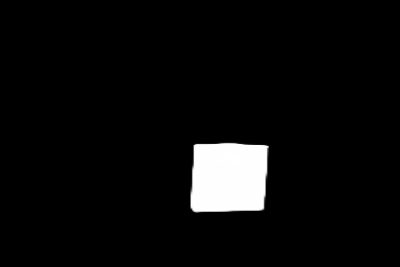}}&
   {\includegraphics[width=0.090\linewidth]{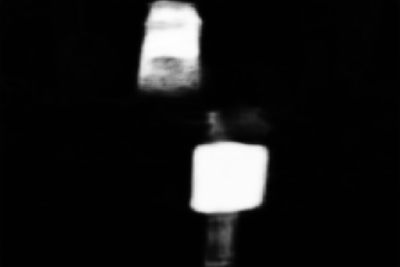}}&
   {\includegraphics[width=0.090\linewidth]{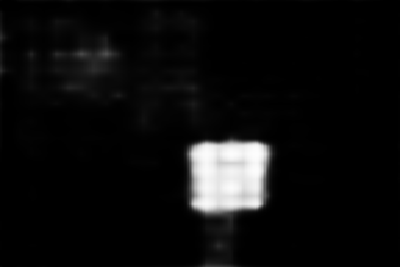}}&
   {\includegraphics[width=0.090\linewidth]{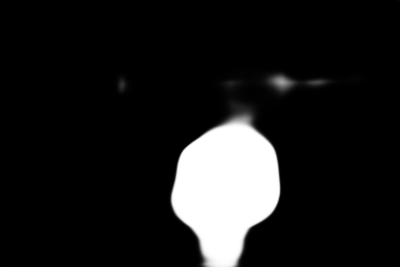}}&
   {\includegraphics[width=0.090\linewidth]{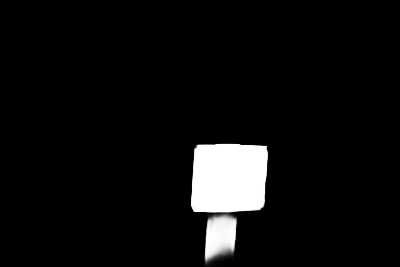}} \\
   {\includegraphics[width=0.090\linewidth]{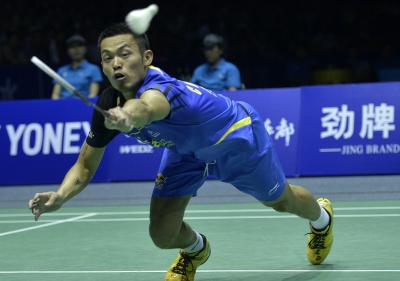}}&
   {\includegraphics[width=0.090\linewidth]{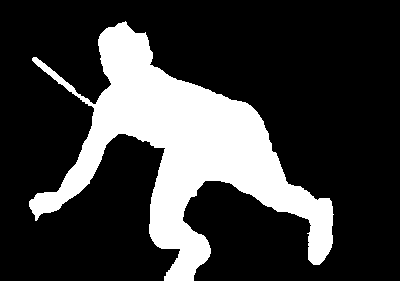}}&
   {\includegraphics[width=0.090\linewidth]{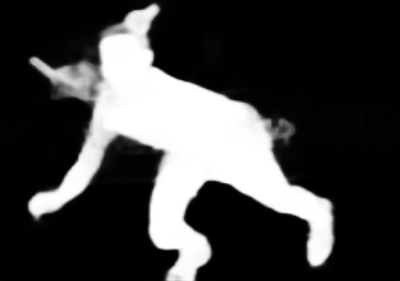}}&
   {\includegraphics[width=0.090\linewidth]{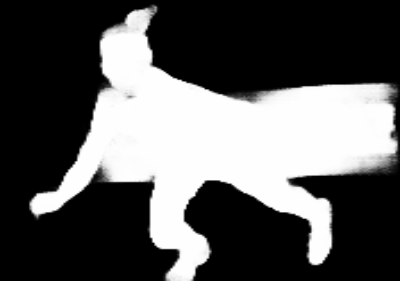}}&
   {\includegraphics[width=0.090\linewidth]{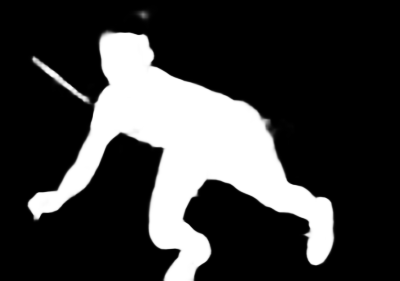}}&
   {\includegraphics[width=0.090\linewidth]{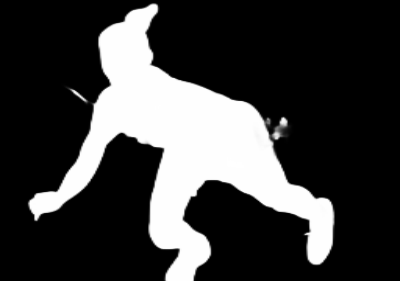}}&
   {\includegraphics[width=0.090\linewidth]{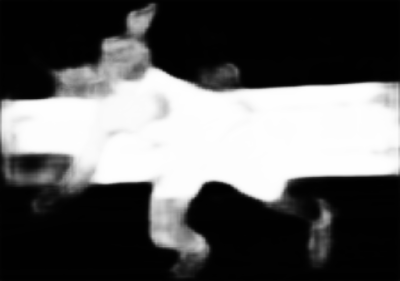}}&
   {\includegraphics[width=0.090\linewidth]{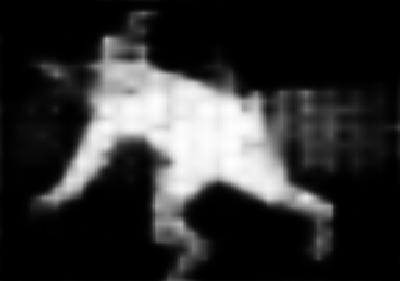}}&
   {\includegraphics[width=0.090\linewidth]{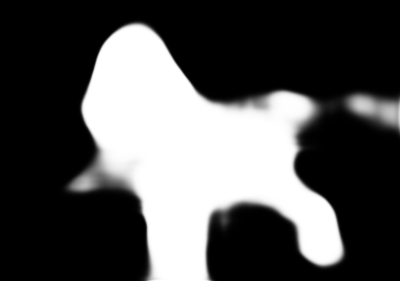}}&
   {\includegraphics[width=0.090\linewidth]{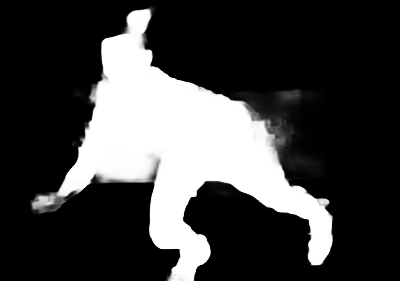}} \\
%   {\includegraphics[width=0.090\linewidth]{sample_show/0002.jpg}}&
%   {\includegraphics[width=0.090\linewidth]{sample_show/0002.png}}&
%   {\includegraphics[width=0.090\linewidth]{sample_show/0002_pica.png}}&
%   {\includegraphics[width=0.090\linewidth]{sample_show/0002_nldf.png}}&
%   {\includegraphics[width=0.090\linewidth]{sample_show/0002_cpd.png}}&
%   {\includegraphics[width=0.090\linewidth]{sample_show/0002_bas.png}}&
%   {\includegraphics[width=0.090\linewidth]{sample_show/0002_sbf.png}}&
%   {\includegraphics[width=0.090\linewidth]{sample_show/0002_msw.png}}&
%   {\includegraphics[width=0.090\linewidth]{sample_show/0002_pce.png}}&
%   {\includegraphics[width=0.090\linewidth]{sample_show/0002_our.png}} \\
%   {\includegraphics[width=0.090\linewidth]{sample_show/0004.jpg}}&
%   {\includegraphics[width=0.090\linewidth]{sample_show/0004.png}}&
%   {\includegraphics[width=0.090\linewidth]{sample_show/0004_pica.png}}&
%   {\includegraphics[width=0.090\linewidth]{sample_show/0004_nldf.png}}&
%   {\includegraphics[width=0.090\linewidth]{sample_show/0004_cpd.png}}&
%   {\includegraphics[width=0.090\linewidth]{sample_show/0004_bas.png}}&
%   {\includegraphics[width=0.090\linewidth]{sample_show/0004_sbf.png}}&
%   {\includegraphics[width=0.090\linewidth]{sample_show/0004_msw.png}}&
%   {\includegraphics[width=0.090\linewidth]{sample_show/0004_pce.png}}&
%   {\includegraphics[width=0.090\linewidth]{sample_show/0004_our.png}} \\
   {\includegraphics[width=0.090\linewidth]{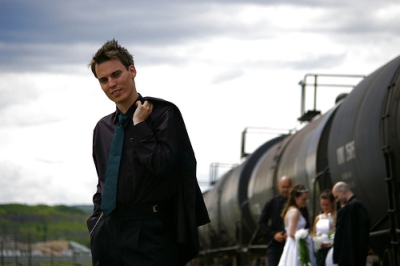}}&
   {\includegraphics[width=0.090\linewidth]{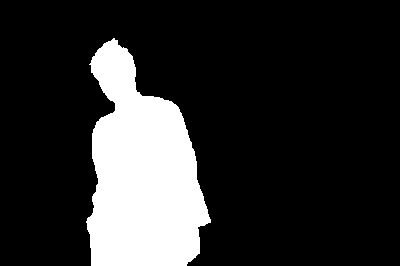}}&
   {\includegraphics[width=0.090\linewidth]{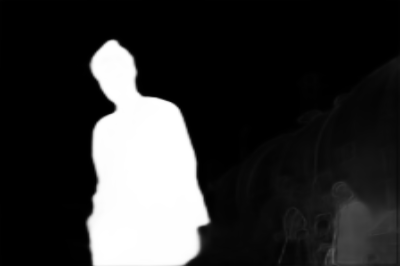}}&
   {\includegraphics[width=0.090\linewidth]{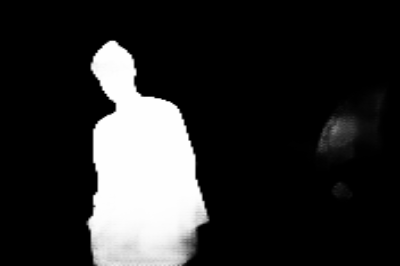}}&
   {\includegraphics[width=0.090\linewidth]{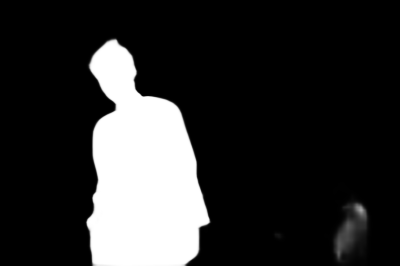}}&
   {\includegraphics[width=0.090\linewidth]{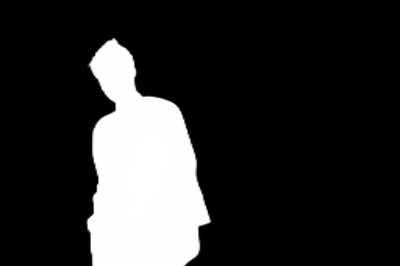}}&
   {\includegraphics[width=0.090\linewidth]{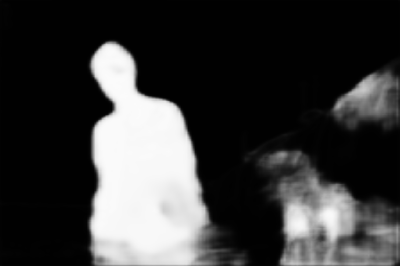}}&
   {\includegraphics[width=0.090\linewidth]{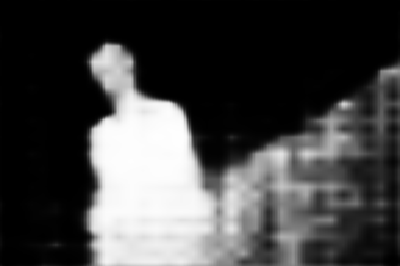}}&
   {\includegraphics[width=0.090\linewidth]{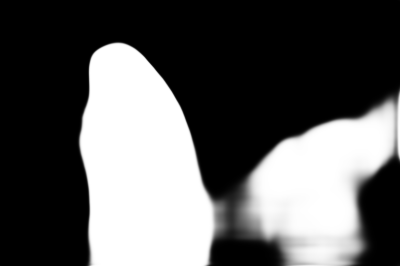}}&
   {\includegraphics[width=0.090\linewidth]{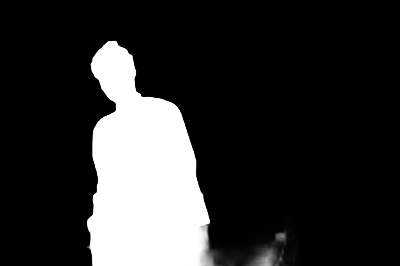}} \\
   {\includegraphics[width=0.090\linewidth]{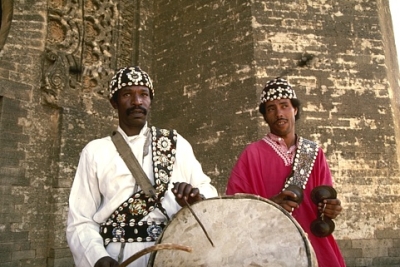}}&
   {\includegraphics[width=0.090\linewidth]{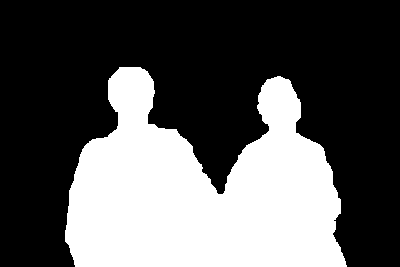}}&
   {\includegraphics[width=0.090\linewidth]{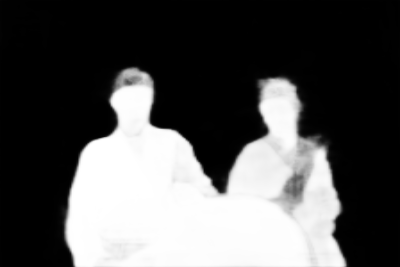}}&
   {\includegraphics[width=0.090\linewidth]{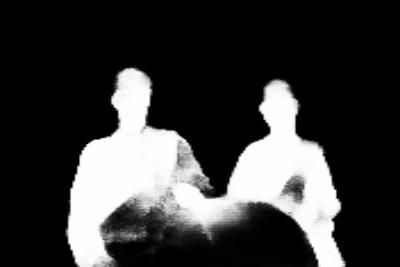}}&
   {\includegraphics[width=0.090\linewidth]{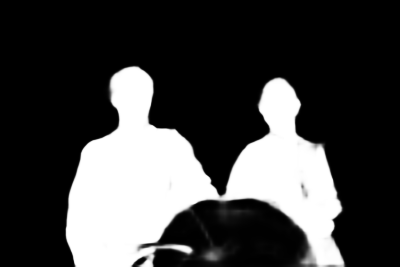}}&
   {\includegraphics[width=0.090\linewidth]{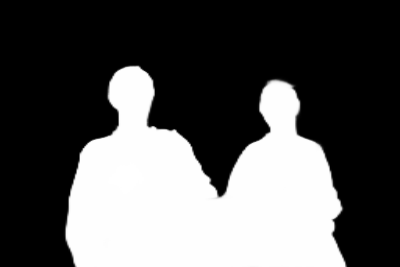}}&
   {\includegraphics[width=0.090\linewidth]{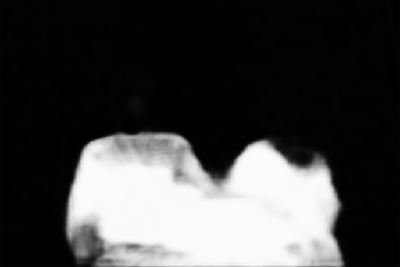}}&
   {\includegraphics[width=0.090\linewidth]{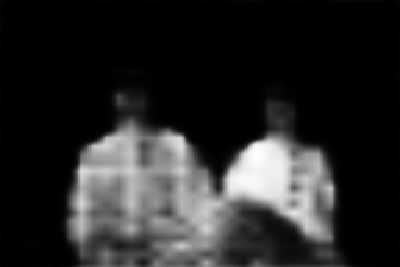}}&
   {\includegraphics[width=0.090\linewidth]{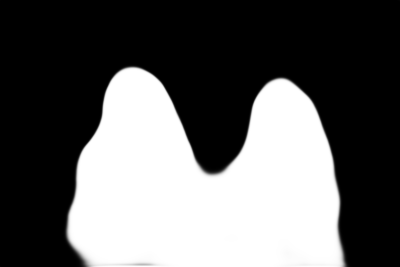}}&
   {\includegraphics[width=0.090\linewidth]{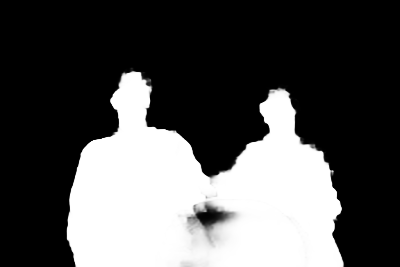}} \\
   \footnotesize{Image} & \footnotesize{GT} & \footnotesize{PiCANet} & \footnotesize{NLDF} & \footnotesize{CPD} & \footnotesize{BASNet} & \footnotesize{SBF} & \footnotesize{MSW} & \footnotesize{M1} & \footnotesize{Ours}\\
   \end{tabular}
   \end{center}
   \vspace{-2mm}
\caption{\small Comparisons of saliency maps. \enquote{M1} represents the results of a baseline model marked as \enquote{M1} in Section \ref{sec_ablation_study}. 
% From left to right: Input image, binary ground-truth, result of PiCANet \cite{picanet}, NLDF \cite{Luo_2017_CVPR}, CPD \cite{CPD_Sal}, BASNet \cite{BASNet_Sal}, SBF \cite{8237698}, MSW \cite{zeng2019Multi}, result of the baseline model "M1" in section \ref{sec_ablation_study} and our final result.
}
   \label{fig:visual_saliency_compare}
   \vspace{-2mm}
\end{figure*}

\subsection{Setup}
\noindent\textbf{Datasets:}
We train our network on our newly labeled scribble saliency dataset: S-DUTS. Then, we evaluate our method on six widely-used benchmarks: (1) DUTS testing dataset \cite{imagesaliency}; (2) ECSSD \cite{Hierarchical:CVPR-2013}; (3) DUT \cite{Manifold-Ranking:CVPR-2013}; (4) PASCAL-S \cite{PASCALS}; (5) HKU-IS \cite{MDF:CVPR-2015} (6) THUR \cite{THUR}.  
% Training: 10,553 scribble-based training images from the proposed S-DUTS dataset.
% Testing: 
% Training dataset: We have evaluated our performance on six saliency benchmarking datasets. We used 10,553 scribble-based training images from the S-DUTS dataset, and 5,019 DUTS testing images for testing. The other testing datasets include: 1) the ECSSD dataset \cite{Hierarchical:CVPR-2013}; 2) the DUT dataset \cite{Manifold-Ranking:CVPR-2013}; 3) the PASCAL-S dataset \cite{PASCALS}; 4) the HKU-IS \cite{MDF:CVPR-2015} dataset and 5) the THUR dataset \cite{THUR}. 

\noindent\textbf{Competing methods:} We compare our method with five state-of-the-art weakly-supervised/unsupervised methods and eleven fully-supervised saliency detection methods.

% : DGRL \cite{Wang_2018_CVPR}, UCF \cite{UCF_ICCV}, PiCANet \cite{picanet}, R3Net \cite{R3Net}, NLDF \cite{Luo_2017_CVPR}, MSNet \cite{MSNet_Sal}, CPD \cite{CPD_Sal}, AFNet \cite{AFNet_Sal},  PFAN \cite{pyramid_attention}, PAGRN \cite{prpgressive_attention}, BASNet \cite{BASNet_Sal}, and five weakly supervised/unsupervised methods: SBF \cite{8237698}, WSI \cite{Guanbin_weaksalAAAI}, WSS \cite{imagesaliency}, MNL \cite{Zhang_2018_CVPR}, MSW \cite{zeng2019Multi}.

% \textbf{Competing methods:} We compared our method against eleven state-of-the-art deep fully-supervised saliency detection methods: DGRL \cite{Wang_2018_CVPR}, UCF \cite{UCF_ICCV}, PiCANet \cite{picanet}, R3Net \cite{R3Net}, NLDF \cite{Luo_2017_CVPR}, MSNet \cite{MSNet_Sal}, CPD \cite{CPD_Sal}, AFNet \cite{AFNet_Sal},  PFAN \cite{pyramid_attention}, PAGRN \cite{prpgressive_attention}, BASNet \cite{BASNet_Sal}, and five weakly supervised/unsupervised methods: SBF \cite{8237698}, WSI \cite{Guanbin_weaksalAAAI}, WSS \cite{imagesaliency}, MNL \cite{Zhang_2018_CVPR}, MSW \cite{zeng2019Multi}.
% , and five conventional methods: DRFI \cite{DRFI:CVPR-2013}, RBD \cite{Background-Detection:CVPR-2014}, DSR \cite{DSR_ICCV13}, MC \cite{MC_ICCV13}, and HS \cite{Hierarchical:CVPR-2013}, which were proven in \cite{SalObjBenchmark_Tip2015} as the state-of-the-art before the deep learning revolution. 

\noindent\textbf{Evaluation Metrics:}
Four evaluation metrics are used,
% to evaluate the performance of our method and the other competing methods, 
including
% two widely used measures (
Mean Absolute Error (MAE $\mathcal{M}$),
% and 
Mean F-measure ($F_{\beta}$),
% one newly proposed metric (
mean E-measure ($E_{\xi}$ \cite{Fan2018Enhanced}) 
and our proposed saliency structure measure ($B_\mu$).
% \textbf{MAE $\mathcal{M}$}: is defined as per-pixel wise difference between predicted
% saliency map $s$ and a per-pixel wise binary ground-truth $y$:
% \begin{equation}
%     \begin{aligned}
%     \text{MAE} = \frac{1}{H\times W}|s-y|,
%     \end{aligned}
% \end{equation}
% where $H$ and $W$ are height and width of $s$. 
% MAE provides a direct
% estimate of conformity between estimated and ground-truth maps.
% However, for the MAE metric, small objects naturally assign a smaller error and
% larger objects have larger errors. The metric also can not tell where the
% error occurs~\cite{tsai2010motion}.
% \textbf{F-measure $F_{\beta}$}: is a region based similarity metric, and we provide the mean F-measure using varying fixed (0-255) thresholds.

% \textbf{E-measure $E_{\xi}$}: is the recent proposed Enhanced alignment
% measure~\cite{Fan2018Enhanced} in the binary map evaluation field to jointly capture image-level statistics and local pixel matching information.

% \textbf{S-measure $S_{\alpha}$}: is a structure based measure ~\cite{fan2017structure}, which combines the region-aware ($S_r$) and object-aware ($S_o$) structural similarity as their final structure metric:
% \begin{equation}
% \label{equ:S-measure}
% S_{\alpha} = \alpha*S_o+(1-\alpha)*S_r,
% \end{equation}
% where $\alpha\!\in\![0,1]$ is the balance parameter and set to 0.5 as default.

\begin{figure*}[!htp]
  \begin{center}
%   {\includegraphics[width=0.24\linewidth]{e_f_curves/ecssd_e.png}}
%   {\includegraphics[width=0.24\linewidth]{e_f_curves/dut_e.png}}
  {\includegraphics[width=0.232\linewidth]{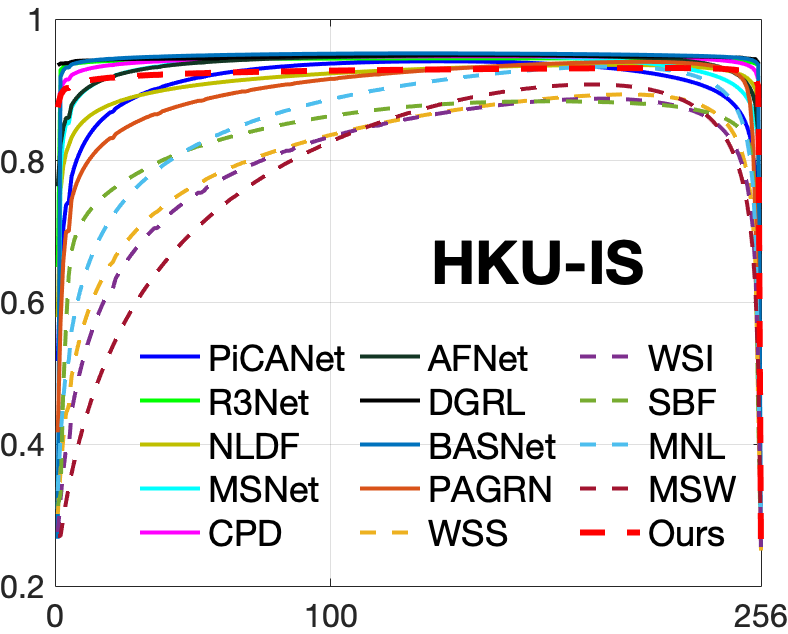}}
  {\includegraphics[width=0.232\linewidth]{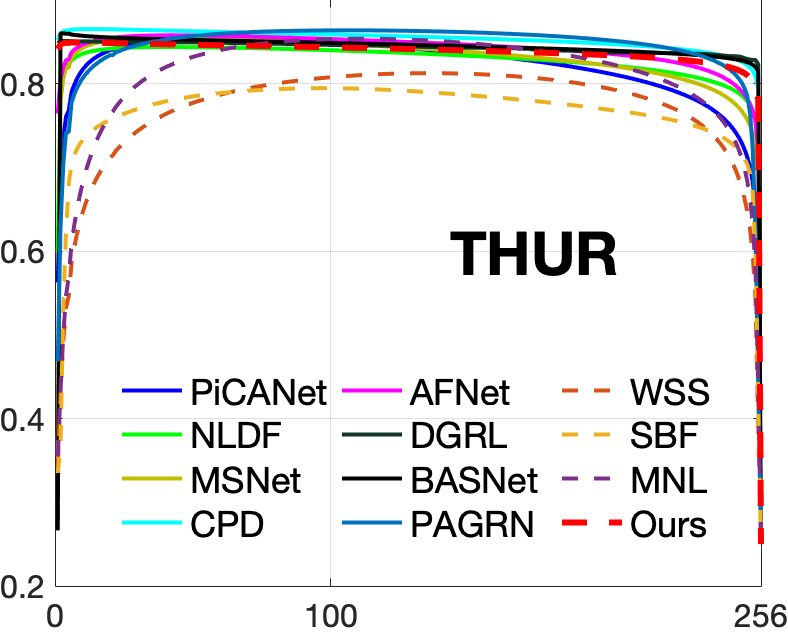}}
%   {\includegraphics[width=0.24\linewidth]{e_f_curves/ecssd_f.png}}
%   {\includegraphics[width=0.24\linewidth]{e_f_curves/dut_f.png}}
  {\includegraphics[width=0.232\linewidth]{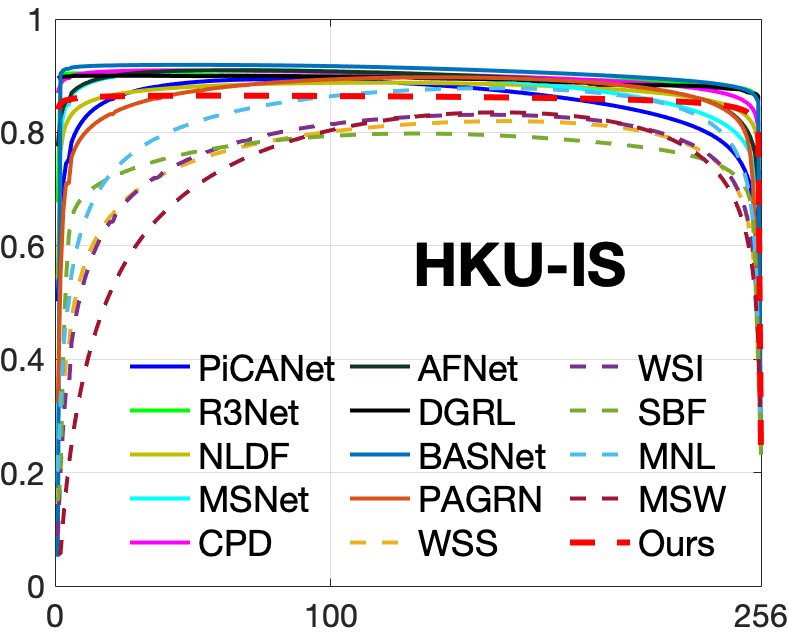}}
  {\includegraphics[width=0.232\linewidth]{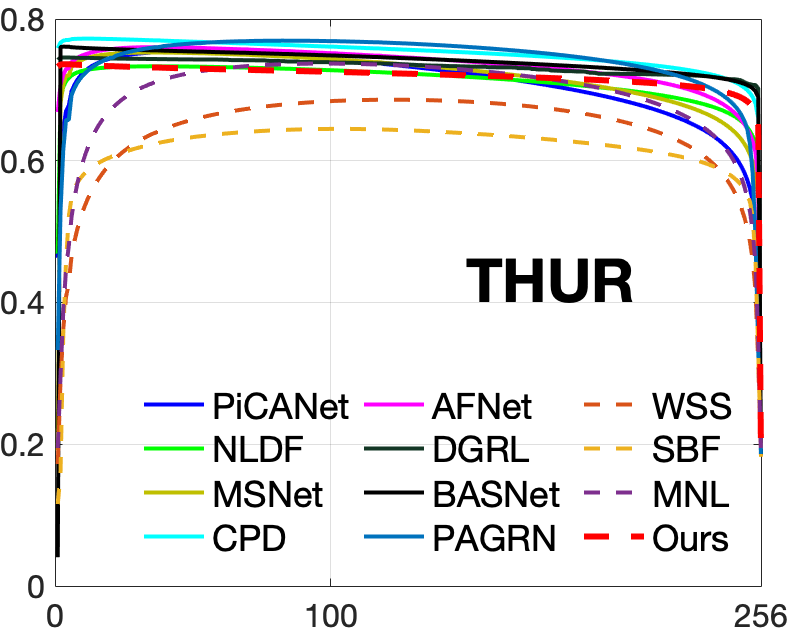}}
  \end{center}
  \vspace{-3mm}
  \caption{E-measure (1$st$ two figures) and F-measure (last two figures) curves on two benchmark datasets.
%   First two figure: E-measure, and last two figures: F-measure.
% %   (HKU-IS and THUR). 
Best Viewed on screen. 
  }
  \vspace{-4mm}
  \label{fig:E_F_measure_show}
\end{figure*}

\subsection{Comparison with the State-of-the-Art}
\noindent\textbf{Quantitative Comparison:}
% We compared performance of the proposed method with eleven deep fully-supervised saliency detection methods and five weakly supervised or unsupervised methods on six benchmark datasets, and show performance in 
% Table \ref{tab:deep_unsuper_Performance_Comparison} and Fig. \ref{fig:E_F_measure_show} show performance of our method compared with competing methods.   
% Especially on the $B_\mu$ measure, those weakly supervised and unsupervised models has no constraints on accuracy of saliency edge, thus they produce high $B_\mu$ measure, while our method achieve much lower $B_\mu$, which is comparable or even outperform some fully-supervised saliency models (DGRL, PiCANet).
% Compared with the other weakly supervised and unsupervised models, our method outperforms 
% which usually produce low E-measure or F-measure with small threshold, our method achieves consistent good performance with varying thresholds, which indicates robustness of our method. 
In Table \ref{tab:deep_unsuper_Performance_Comparison} and Fig.~\ref{fig:E_F_measure_show}, we compare our
% quantitative 
results with other competing methods. As indicated in Table \ref{tab:deep_unsuper_Performance_Comparison}, we achieves consistently the best performance compared with other weakly-supervised or unsupervised methods under these four saliency evaluation metrics.
Since state-of-the-art weakly-supervised or unsupervised models do not impose any constraints on the boundaries of predicted saliency maps, these methods cannot preserve the structure in the prediction and produce high values on $B_\mu$ measure. In contrast, our method explicitly enforces a gated structure-aware loss to the edges of the prediction, and achieves lower $B_\mu$. Moreover, our performance is also comparable or superior to some fully-supervised saliency models, such as DGRL and PiCANet.
Fig. \ref{fig:E_F_measure_show} shows the E-measure and F-measure curves of our method as well as the other competing methods on HKU-IS and THUR datasets. Due to limits of space, E-measure and F-measure curves on the other four testing datasets are provided in the supplementary material.
As illustrated in Fig. \ref{fig:E_F_measure_show}, our method significantly outperforms the other weakly-supervised and unsupervised models with different thresholds, demonstrating the robustness of our method. Furthermore, the performance of our method is also on par with some fully-supervised methods as seen in Fig. \ref{fig:E_F_measure_show}.

\noindent\textbf{Qualitative Comparison:} We sample four images from the ECSSD dataset \cite{Hierarchical:CVPR-2013} and the saliency maps predicted by six competing methods and our method are illustrated in Fig.~\ref{fig:visual_saliency_compare}. Our method, while achieving performance on par with some fully-supervised methods, significantly outperforms other weakly-supervised and unsupervised models.
In Fig.~\ref{fig:visual_saliency_compare}, we further show that directly training with scribbles produces saliency maps with poor localization (\enquote{M1}).
% With the proposed edge-detection task and boundary-aware loss to highlight object boundaries and penalize prediction of inconsistent boundary with RGB image, our method produce relatively sharp saliency map.
Benefiting from our EDN as well as gated structure-aware loss, our network is able to produce sharper saliency maps than other weakly-supervised and unsupervised ones.

\subsection{Ablation Study}
\label{sec_ablation_study}

\begin{table}[t!]
  \centering
  \scriptsize
  \renewcommand{\arraystretch}{1.1}
  \renewcommand{\tabcolsep}{1.3mm}
  \caption{Ablation study on six benchmark datasets.
%   \XY{7 or 6?}
%   $\uparrow \& \downarrow$ denote larger and smaller is better, respectively.
  }
  \begin{tabular}{lr|cccccccccc}
  \hline
%   \toprule
    & Metric &M0&
   M1  & M2  &M3  & M4    & M5 &
   M6 & M7&M8 &M9\\
  \hline
  \multirow{4}{*}{\begin{sideways}\textit{ECSSD}\end{sideways}}
    & $B_{\mu}\downarrow$    & .550& .896 & .592 & .616& .714 & .582 & .554 & .771 & .543 &  .592  \\
    & $F_{\beta}\uparrow$     & .865& .699 & .823& .804 & .778 & .845 & .835 & .696 & .868 & .839\\
    & $E_{\xi}\uparrow$       & .908& .814 & .874& .859 & .865 & .898 & .890 & .730 & .908 &  .907\\
    & $\mathcal{M}\downarrow$ & .061& .117 & .083 & .094& .091 & .068 & .074 & .136 & .059 &  . 070 \\
    \hline
    \multirow{4}{*}{\begin{sideways}\textit{DUT}\end{sideways}}
    & $B_{\mu}\downarrow$    & .655& .925 & .696 & .711& .777 & .685 & .665 & .786 & .656 &  .708  \\
    & $F_{\beta}\uparrow$     & .702& .518 & .656 & .626& .580 & .679 & .658 & .556 & .691 &  .671  \\
    & $E_{\xi}\uparrow$       & .835& .699 & .807 & .774& .743 & .823 & .805 & .711 & .823 &  .816 \\
    & $\mathcal{M}\downarrow$ & .068& .134 & .083 & .102& .116 & .074 & .081 & .108 & .069 &  .080  \\
   \hline
    \multirow{4}{*}{\begin{sideways}\textit{PASCAL-S}\end{sideways}}
    & $B_{\mu}\downarrow$    & .665& .921 & .732 & .760& .787 & .693 & .676 & .792 & .664 &  .722 \\
    & $F_{\beta}\uparrow$    & .788 & .693 & .748 & .727& .741 & .772 & .768 & .657 & .792 & .771  \\
    & $E_{\xi}\uparrow$      & .798 & .761 & .757 & .731& .795 & .791 & .782 & .664 & .800 &  .804 \\
    & $\mathcal{M}\downarrow$ & .140& .171 & .160 & .173& .152 & .145 & .152 & .204 & .136 & .143 \\
    \hline
    \multirow{4}{*}{\begin{sideways}\textit{HKU-IS}\end{sideways}}
    & $B_{\mu}\downarrow$    & .537& .892 & .567 & .609& .670 & .574 & .559 & .747 & .535 & .564 \\
    & $F_{\beta}\uparrow$     & .858& .651 & .813 & .789& .747 & .835 & .812 & .646 & .857 & .821\\
    & $E_{\xi}\uparrow$      & .923 & .799 & .904 & .878& .867 & .911 & .900 & .761 & .920 &  .907 \\
    & $\mathcal{M}\downarrow$& .047 & .113 & .060 & .083& .080 & .055 & .062 & .123 & .047 &  .058  \\
    \hline
    \multirow{4}{*}{\begin{sideways}\textit{THUR}\end{sideways}}
    & $B_{\mu}\downarrow$   & .596 & .927 & .637 & .677& .751 & .635 & .606 & .780 & .592 & .650 \\
    & $F_{\beta}\uparrow$    & .718 & .520 & .660 & .641& .596 & .696 & .683 & .586 & .718 & .690 \\
    & $E_{\xi}\uparrow$    & .837   & .687 & .803 & .773& .750 & .824 & .814 & .718 & .834 &  .804  \\
    & $\mathcal{M}\downarrow$& .077 & .150 & .099 & .118& .123 & .085 & .087 & .125 & .078 & .086 \\
    \hline
  \multirow{4}{*}{\begin{sideways}\textit{DUTS}\end{sideways}}
    & $B_{\mu}\downarrow$  & .603  & .923 & .681 & .708& .763 & .639 & .634 & .745 & .604 & .687 \\
    & $F_{\beta}\uparrow$  & .747   & .517 & .688 & .652& .607 & .728  & .685 & .578 & .743 & .728  \\
    & $E_{\xi}\uparrow$   & .865    & .699 & .833 & .805& .776 & .857 & .828 & .719 & .856 & .855 \\
    & $\mathcal{M}\downarrow$& .062 & .135 & .079 & .101& .106 & .068 & .080 & .106 & .061 & .080 \\
    % \bottomrule
  \hline
  \end{tabular}
  \vspace{-4mm}
  \label{tab:ablation_study}
\end{table}

We carry out nine experiments (as shown in Table \ref{tab:ablation_study}) to analyze our method, including our loss functions (\enquote{M1}, \enquote{M2} and \enquote{M3}), network structure (\enquote{M4}), DenseCRF post-processing (\enquote{M5}), scribble boosting strategy (\enquote{M6}), scribble enlargement (\enquote{M7}) and robustness analysis (\enquote{M8}, \enquote{M9}). Our final result is denoted as \enquote{M0}.
% We thoroughly an
% alyze our proposed method and show performance in Table \ref{tab:ablation_study}, where \enquote{M0} is our final performance.

\noindent\textbf{Direct training with scribble annotations:} 
% A direct solution is to learn saliency from provided scribble annotations.
We employ the partial cross-entropy loss to train our SPN in Fig. \ref{fig:overview} with scribble labels. The performance is marked as \enquote{M1}.
% We remove our edge detection branch and the gated structure-aware loss function and then train the network directly on scribbles, and the performance is shown in \enquote{M1}, 
% which is much worse than out result in \enquote{M0},
% % , we find that directly training with the scribble data fails to achieve good performance, 
% especially, \enquote{M1} leads to high $B_\mu$ measure, indicating poor saliency map localization ability as shown in Fig. \ref{fig:figure1} \textcolor{red}{(e)}.
As expected, \enquote{M1} is much worse than our result \enquote{M0} and the high $B_\mu$ measure also indicates that object structure is not well preserved if only using the partial cross-entropy loss.

\noindent\textbf{Impact of gated structure-aware loss:} 
We add our gated structure-aware loss to \enquote{M1}, and the performance is denoted by \enquote{M2}. 
The gated structure-aware loss improves the performance in comparison with \enquote{M1}. However, without using our EDN, \enquote{M2} is still inferior to \enquote{M0}.

\noindent\textbf{Impact of gate:}
We propose gated structure-aware loss to let the network focus on salient regions of images instead of the entire image as in the traditional smoothness loss \cite{occlusion_aware}. 
To verify the importance of the gate, we compare our loss with the smoothness loss, marked as \enquote{M3}. As indicated, \enquote{M2} achieves better performance than \enquote{M3}, demonstrating the gate reduces the ambiguity of structure recovery.
% To verify effectiveness of the new loss, we define smoothness loss (with the same hyper-parameter $\alpha$) as $\mathcal{L}_b$ in Eq. \ref{final_loss}, and show performance in \enquote{M3}.

% We remove both edge-detection module and structure-aware loss from our framework, and obtain poor performance as shown in \enquote{M1}.
% We discard the edge detection branch (as well as the edge-enhanced module) to test the effectiveness of the edge-detection task, and show performance in \enquote{M4}, which is worse than our results in \enquote{M0}, thus effectiveness of edge detection task can be confirmed.
 
\noindent\textbf{Impact of the edge detection task:}
We add edge detection task to \enquote{M1}, and use cross-entropy loss to train the EDN. Performance is indicated by \enquote{M4}.
% which is clearly better than \enquote{M1}. 
We observe that the $B_\mu$ measure is significantly decreased compared to \enquote{M1}. This indicates that our auxiliary edge-detection network provides rich structure guidance for saliency prediction. Note that, our gated structure-aware loss is not used in \enquote{M4}.

\noindent\textbf{Impact of scribble boosting:}
We employ all the branches as well as our proposed losses to train our network and the performance is denoted by \enquote{M5}. The predicted saliency map is also called our initial estimated saliency map. We observe decreased performance compared with \enquote{M0}, where one iteration of scribble boosting is employed, which indicates effectiveness of the proposed boosting scheme.
% Compared with the results \enquote{M5}, our final results \enquote{M0}, of which we employ our scribble boosting scheme, achieves better saliency detection performance.

% We introduced scribble boosting (\cf Section 3.2) to use DenseCRF as post-processing to obtain expanded scribble annotation, which is then adopt as supervision signal for the next iteration training. Performance in \enquote{M6} indicates the initial performance before we boost the scribble. We observe inferior performance in \enquote{M6} as compared with \enquote{M0}, while \enquote{M6} consistently outperforms existing weakly supervised methods. 

\noindent\textbf{Employing DenseCRF as post-processing:} 
After obtaining our initial predicted saliency map, we can also use post-processing techniques to enhance the boundaries of the saliency maps. Therefore, we refine \enquote{M5} with DenseCRF, and results are shown in \enquote{M6}, which is inferior to \enquote{M5}. The reason lies in two parts: 1) the hyperparameters for DenseCRF is not the best; 2) DenseCRF recover structure information without considering saliency of the structure, causing extra false positive region.
% we exploit DenseCRF as our post-processing method to refine \enquote{M5}. The results processed by DenseCRF are marked as \enquote{M6}.
% As expected, DenseCRF is able to improve the structure of our initial saliency maps, attaining lower $B_\mu$ measure than \enquote{M5}. 
Using our scribble boosting mechanism, we can always achieve boosted or at least comparable performance as indicated by \enquote{M0}.
\noindent\textbf{Using Grabcut to generate pseudo label:} Given scribble annotation, one can enlarge the annotation by using
% interactive segmentation tools, \eg 
Grabcut \cite{rother2004grabcut}. We carried out experiment with pseudo label $y'$ obtained by applying Grabcut to our scribble annotations $y$, and show performance in \enquote{M7}. During training, we employ the same loss function as in Eq. \ref{final_loss}, except that we use cross-entropy loss for $\mathcal{L}_s$. Performance of \enquote{M7} is worse than ours. The main reason is that pseudo label $y'$ contains noise due to limited accuracy of Grabcut. Training directly with $y'$ will overwhelm the network remembering the noisy label instead of learning useful saliency information.

\noindent\textbf{Robustness to different scribble annotations:} 
% We re-labeled three versions of the DUTS dataset by three different annotators.
% To test how our model performs with different scribble dataset, we trained our model with a different version of S-DUTS dataset, and the performance is shown as \enquote{M7}, which is comparable with the reported results in \enquote{M0}. Results of \enquote{M7} indicates the robustness of our proposed model to different scribble annotations. 
We report our performance \enquote{M0} by training the network with one set of scribble dataset. We then train with another set of the scribble dataset (\enquote{M8}) to test robustness of our model.
% In training our network, we only select one set of scribble annotations provided by one labeler and the testing performance is \enquote{M0}. 
% To test robustness of our model to different scribbles, we train network with another set of scribble annotations,
% We choose another scribble annotations labeled by another labeler to train our network 
% and show results in \enquote{M8}. 
% As seen in Table \ref{tab:ablation_study}, 
We observe staple performance compared with
% the performance is stable compared with
\enquote{M0}. This implies that our method is robust to the scribble annotations despite their sparsity and few overlaps annotated by different labelers. 
% Moreover, 
We also conduct experiments with merged scribbles of different labelers as supervision signal and show performance of this experiment
% by randomly choosing annotations provided by different labelers as well as merging annotated scribbles as supervision for training our network. Due to the space limit, we demonstrate those results 
in the supplementary material.
%%% \YD{It seems M6 is consistently inferior to M0 even though the gap is very small. Any reason?}

% respectively, which are comparable with the reported results in Table \ref{tab:deep_unsuper_Performance_Comparison}. Results of \enquote{M6} and \enquote{M7} indicate robustness of the proposed model to different scribble annotations.

\noindent\textbf{Different edge detection methods:} 
We obtain the edge maps $E$ in Eq. \ref{edge_loss} from RCF edge detection network \cite{ChengEdge} to train EDN. 
We also employ a hand-crafted edge map detection method, \enquote{Sobel}, to train EDN, denoted by \enquote{M9}. Since Sobel operator is more sensitive to image noise compared to RCF, \enquote{M9} is a little inferior to \enquote{M0}. However, \enquote{M9} still achieves better performance than the results without using EDN, such as \enquote{M1}, \enquote{M2} and \enquote{M3}, which further indicates effectiveness of the edge detection module.

\vspace{-2mm}
\section{Conclusions}
In this paper, we proposed a weakly-supervised salient object detection (SOD) network trained on our newly labeled scribble dataset (S-DUTS). Our method significantly relaxes the requirement of labeled data for training a SOD network. By introducing an auxiliary edge detection task and a gated structure-aware loss, our method produces saliency maps with rich structure, which is more consistent with human perception measured by our proposed saliency structure measure. Moreover, we develop a scribble boosting mechanism to further enrich scribble labels.
% , and thus achieve better salient object detection performance. 
Extensive experiments demonstrate that our method significantly outperforms state-of-the-art weakly-supervised or unsupervised methods and is on par with fully-supervised methods.

\vspace{1mm}
\noindent\textbf{Acknowledgment.}
This research was supported in part by Natural  Science  Foundation  of  China  grants  (61871325, 61420106007, 61671387), the Australia Research Council Centre of Excellence for Robotics Vision (CE140100016), and the National Key Research and Development Program of China under Grant 2018AAA0102803. We thank all reviewers and Area Chairs for their constructive comments.

{\small
\bibliographystyle{ieee_fullname}
\bibliography{Scribble_Saliency_Arxiv}
}

\end{document}